\documentclass[journal]{IEEEtran}

\usepackage{epsfig}
\usepackage{amsmath}
\usepackage{amssymb}
\usepackage{newalg}
\usepackage{float}
\floatstyle{ruled}
\newfloat{model}{thp}{lop}
\floatname{model}{Model}

\usepackage{color}

\begin{document}
\title{A Linear-Programming Approximation of \\ AC Power Flows}

\author{
	Carleton~Coffrin,~\IEEEmembership{Member,~IEEE}, 
	Pascal~Van~Hentenryck,~\IEEEmembership{Member,~IEEE}

        \thanks{C. Coffrin and P. Van Hentenryck are members of the
          Optimization Research Group, NICTA, Victoria 3010,
          Australia. P. Van Hentenryck is Professor in the School of
          Engineering at the University of Melbourne.}}

\maketitle

\begin{abstract}
  Linear active-power-only DC power flow approximations are pervasive
  in the planning and control of power systems.  However, these
  approximations fail to capture reactive power and voltage
  magnitudes, both of which are necessary in many applications to
  ensure voltage stability and AC power flow feasibility.  This paper
  proposes linear-programming models (the LPAC models) that
  incorporate reactive power and voltage magnitudes in a linear power
  flow approximation.  The LPAC models are built on a convex
  approximation of the cosine terms in the AC equations, as well as
  Taylor approximations of the remaining nonlinear terms. Experimental
  comparisons with AC solutions on a variety of standard IEEE and 
  MATPOWER benchmarks show that the LPAC models produce accurate
  values for active and reactive power, phase angles, and voltage
  magnitudes.  The potential benefits of the LPAC models are
  illustrated on two ``proof-of-concept'' studies in power restoration
  and capacitor placement.
\end{abstract}

\begin{IEEEkeywords}
DC power flow, AC power flow, LP power flow, linear relaxation, power system analysis, capacitor placement, power system restoration
\end{IEEEkeywords}

\IEEEpeerreviewmaketitle

\vspace{-0.2cm}
\section*{Nomenclature}
\addcontentsline{toc}{section}{Nomenclature}
\begin{IEEEdescription}[\IEEEusemathlabelsep\IEEEsetlabelwidth{$V_1,V_2,V_3,V_4,$}]
\item[$\widetilde{I}$] AC Current
\item[$\widetilde{V} = v+i\theta$] AC voltage
\item[$\widetilde{S} = p+iq$] AC power
\item[$\widetilde{Z} = r+ix$] Line impedance
\item[$\widetilde{Y} = g+ib$] Line admittance
\item[$\widetilde{Y}^b = g^y+ib^y$] Y-Bus element
\item[$\widetilde{Y}^c = g^c+ib^c$] Line charge
\item[$\widetilde{Y}^s = g^s+ib^s$] Bus shunt
\item[$\widetilde{T} = t+is$] Transformer parameters
\vspace{0.1cm}
\item[$\widetilde{V} = |\widetilde{V}| \angle \theta^\circ$] Polar form
\vspace{0.1cm}
\item[$\widetilde{S}_n$] AC Power at bus $n$
\item[$\widetilde{S}_{nm}$] AC Power on a line from $n $ to $m$
\item[${\cal PN}$] Power network
\item[$N$] Set of buses in a power network
\item[$L$] Set of lines in a power network
\item[$G$] Set of voltage controlled buses
\item[$s$] Slack Bus
\item[$|\widetilde{V}^h|$] Hot-Start voltage magnitude
\item[$|\widetilde{V}^t|$] Target voltage magnitude
\item[$\phi$] Voltage magnitude change
\item[$\Delta$] Absolute difference
\item[$\delta$] Percent difference
\item[$\hat{x}$] Approximation of $x$
\item[$\overline{x}$] Upper bound of $x$
\item[$\underline{x}$] Lower bound of $x$
\end{IEEEdescription}

\newpage

\section{Introduction}
\IEEEPARstart{O}{ptimization} 
technology is widely used in modern power systems
\cite{9780824791056} and has resulted in millions of dollars in
savings annually \cite{PJMStudy}. But the increasing role of demand
response, the integration of renewable sources of energy, and the
desire for more automation in fault detection and recovery pose new
challenges for the planning and control of electrical power
systems \cite{SmatyGridStudy}. Power grids now need to operate in more stochastic
environments and under varying operating conditions, while still
ensuring system reliability and security.

Optimization of power systems encompasses a broad spectrum of problem
domains, including optimal power flow
\cite{Wang1996153,IJEEE2010,Thukaram198472,5971792,317548,5712480,54549,317656,192899,744492,744495},
LMP-base market calculations
\cite{Overbye:2004vb,IJIESP2009,Rafael_analysisand}, transmission
switching \cite{4492805,4957010,5460912}, real-time
security-constrained dispatch \cite{4075530,720058}, day-ahead
security-constrained unit commitment
\cite{317600,9780471443377,4076005}, distribution network
configuration \cite{Borghetti:2011pscc,RomeroRamos2010562}, capacitor
placement \cite{Aguiar:2005pscc,871731,544656}, expansion planning
\cite{Bienstock2007115,Mangoli199341,4110854,651620,DBLP:conf/inoc/KosterL11,Jabr2008941,9780077074241,32462,5772044},
vulnerability analysis
\cite{1294998,4762170,4074959,4402321,DBLP:journals/siamjo/BienstockV10},
and power system restoration \cite{SSP1,PRVRP1} to name a few.  Some
of these use active power only,
while others consider both active and reactive power.

Restricting attention to active power is often appealing
computationally as the nonlinear AC power flow equations can then be
approximated by a set of linear equations that define the so-called
Linearized DC (LDC) model. Under normal operating conditions and with
some adjustment for line losses, the LDC model produces a reasonably
accurate approximation of the AC power flow equations for active power
\cite{Stott:2009bb}. Moreover, the LDC model can be embedded in
Mixed-Integer Programming (MIP) models for a variety of optimization
applications in power system operations. This is particularly
attractive as the computational efficiency of Linear Programming (LP)
and MIP solvers has significantly improved over the last two decades
\cite{Bixby2000}.


However, the LDC model does not capture reactive power and hence
cannot be used for applications such as capacitor placement and
voltage stability to name only two. Moreover, the accuracy of the LDC
model outside normal operating conditions is an open point of
discussion (e.g.,
\cite{Overbye:2004vb,Stott:2009bb,Purchala:2005gt,PES1,PES2}).  This
in turn raises concerns for other applications such transmission
planning, vulnerability analysis, and power restoration, which may return infeasible or
suboptimal solutions when the LDC model is used to approximate the AC
power flow equations. As a result, these applications often turn to
nonlinear programming techniques
\cite{317548,744492,744495,Rafael_analysisand}, iterative heuristics
and decomposition \cite{317656,192899,Jabr2008941,4074959}, model
relaxation \cite{5971792,Aguiar:2005pscc}, tabu search \cite{544656},
and genetic algorithms \cite{871731,651620} to ensure feasibility.
These techniques often require extensive tuning for each problem
domain, may consume significant computational resources, and cannot
guarantee global optimality.

This paper aims at bridging the gap between the LDC model and the AC
power flow equations. It presents linear programs to approximate the
AC power flow equations. These linear programs, called the LPAC models,
are based on two ideas:
\begin{enumerate}
\item They reason both on the voltage phase angles and the voltage
  magnitudes, which are coupled through equations for active and
  reactive power;

\item They use a piecewise linear approximation of the cosine term in
  the power flow equations and Taylor series for approximating the
  remaining nonlinear terms.
\end{enumerate}

\noindent
The LPAC models have been evaluated experimentally over a number of
standard benchmarks under normal operating conditions and various
contingencies. Experimental comparisons with AC solutions on standard
IEEE and MatPower benchmarks shows that the LPAC models are highly
accurate for active and reactive power, phase angles, and voltage
magnitudes. Moreover, the LPAC models can be integrated in MIP models
for applications reasoning about reactive power (e.g., capacitor
placement) or topological changes (e.g., transmission planning,
vulnerability analysis, and power restoration).

This rest of this paper presents a rigorous and systematic derivation
of the LPAC models, experimental results about their accuracy, and its
application to power restoration and capacitor placement. Section
\ref{section:review} reviews the AC power flow equations. Section
\ref{section:LAC} derives the LPAC models and Section
\ref{section:Accuracy} presents the experimental results on its
accuracy. Section \ref{section:casestudies} presents the
``proof-of-concept" experiments in power restoration and capacitor
placement to demonstrate potential applications of the LPAC models. 
Section \ref{section:Related} discusses related work and Section
\ref{section:conclusion} concludes the paper.

\section{Review of AC Power Flow}
\label{section:review}


The steady state AC power for bus $n$ is given by
\begin{eqnarray}
\widetilde{S}_n = \sum_m^{n \neq m} \widetilde{V}_{n}\widetilde{V}_{n}^*\widetilde{Y}_{nm}^* - \widetilde{V}_{n}\widetilde{V}_{m}^*\widetilde{Y}_{nm}^*. \label{eq:complexPowerFlow}
\end{eqnarray}
This equation is not symmetric. From the perspective of bus $n$, the
power flow on a line to bus $n$ is
\[
\widetilde{V}_{n}\widetilde{V}_{n}^*\widetilde{Y}_{nm}^* - \widetilde{V}_{n}\widetilde{V}_{m}^*\widetilde{Y}_{nm}^*
\]
while, from the perspective of bus $m$, it is
\[
\widetilde{V}_{m}\widetilde{V}_{m}^*\widetilde{Y}_{mn}^* - \widetilde{V}_{m}\widetilde{V}_{n}^*\widetilde{Y}_{mn}^*.
\]
In general, $\widetilde{S}_{nm} \neq \widetilde{S}_{mn}$.

\subsection{The Traditional Representation}

The AC power flow definition is typically expanded in terms of real
numbers only. By representing power in rectangular form, the real
($p_n$) and imaginary ($q_n$) terms become
\begin{small}
\begin{eqnarray*}
\sum_m^{n \neq  m} \!|\widetilde{V}_n|^{2}g_{nm} - |\widetilde{V}_n||\widetilde{V}_m|(g_{nm}\cos(\theta^\circ_n \!\!-\! \theta^\circ_m) \!\!+\! b_{nm}\sin(\theta^\circ_n \!\!-\! \theta^\circ_m)) \\
\sum_m^{n \neq  m} \!-|\widetilde{V}_n|^{2}b_{nm} \!- |\widetilde{V}_n||\widetilde{V}_m|(g_{nm}\sin(\theta^\circ_n \!\!-\! \theta^\circ_m) \!\!-\!  b_{nm}\cos(\theta^\circ_n \!\!-\! \theta^\circ_m)) 
\end{eqnarray*}
\end{small}
{\em The Y-Bus Matrix: } The formulation can be simplified further by
using a {\em Y-Bus Matrix}, i.e., a precomputed lookup table that
allows the power flow at each bus to be written as a summation of $2n$
terms instead a summation of $3(n-1)$ terms.  Observe that, in the
power flow equations (\ref{eq:complexPowerFlow}), the first term
$\widetilde{V}_{n}\widetilde{V}_{n}^*\widetilde{Y}_{nm}^*$ is a
special case of the second term
$-\widetilde{V}_{n}\widetilde{V}_{m}^*\widetilde{Y}_{nm}^*$ with
$-\widetilde{V}_{m} = \widetilde{V}_{n}$. We can eliminate this
special case by
\begin{enumerate}
\item extending the summation to include $n$ terms,
i.e., $\sum_m^{n \neq m}$ becomes $\sum_m$; 
\item defining the {\em Y-Bus} admittance $\widetilde{Y}_{nm}^b$ as
\begin{eqnarray}
\widetilde{Y}_{nn}^b &=& \sum_m^{n \neq m} \nonumber \widetilde{Y}_{nm}\\
\widetilde{Y}_{nm}^b &=& -\widetilde{Y}_{nm} \nonumber 
\end{eqnarray}
\end{enumerate}
Given the Y-Bus, the power flow equations (\ref{eq:complexPowerFlow})
can be rewritten as a single summation
\begin{eqnarray}
\widetilde{S}_n = \sum_m \widetilde{V}_{n}\widetilde{V}_{m}^*\widetilde{Y}_{nm}^{b} \label{eq:complexPowerFlowYBus}
\end{eqnarray}
giving us the popular formulation of active and reactive power:
\begin{eqnarray}
p_n &=& \!\sum_m \!|\widetilde{V}_n||\widetilde{V}_m|(g_{nm}^y\cos(\theta^\circ_n \!\!-\! \theta^\circ_m) \!+\! b_{nm}^y\sin(\theta^\circ_n \!\!-\! \theta^\circ_m)) \label{eq:acActiveYbus} \\
q_n &=& \!\sum_m \!|\widetilde{V}_n||\widetilde{V}_m|(g_{nm}^y\sin(\theta^\circ_n \!\!-\! \theta^\circ_m) \!-\!  b_{nm}^y\cos(\theta^\circ_n\!\!-\! \theta^\circ_m))  \label{eq:acReactiveYbus} 
\end{eqnarray}

\subsection{An Alternate Representation}
\label{section:powerflow}

The Y-Bus formulation is concise but makes it difficult to reason
about the power flow equations. This paper uses the more explicit
equations which can be presented as bus and line equations as
follows:
\begin{eqnarray}
p_n &=& \sum_m^{n \neq  m} p_{nm} \label{eq:altAC1}\\
q_n &=& \sum_m^{n \neq  m} q_{nm} \\
p_{nm} &=& |\widetilde{V}_n|^{2}g_{nm} - |\widetilde{V}_n||\widetilde{V}_m|g_{nm}\cos(\theta^\circ_n - \theta^\circ_m) \nonumber \\
              &  & - |\widetilde{V}_n||\widetilde{V}_m|b_{nm}\sin(\theta^\circ_n - \theta^\circ_m)  \label{eq:acActiveLine} \\
q_{nm} &=&  -|\widetilde{V}_n|^{2}b_{nm} + |\widetilde{V}_n||\widetilde{V}_m|b_{nm}\cos(\theta^\circ_n - \theta^\circ_m) \nonumber \\ 
               &  & -|\widetilde{V}_n||\widetilde{V}_m|g_{nm}\sin(\theta^\circ_n - \theta^\circ_m) \label{eq:acReactiveLine}  \label{eq:altAC4}
\end{eqnarray}
Once again, \ref{eq:acActiveLine} and \ref{eq:acReactiveLine} are
asymmetric and the line admittance values $\widetilde{Y}$ have not been modified.

\subsection{Extensions for Practical Power Networks}
\label{Section:particalnetworks}

In the above derivation, each line is a conductor with an impedance
$\widetilde{Z}$. The formulation can be extended to line charging and
other components such as transformers and bus shunts, which are
present in nearly all AC system benchmarks. We show how to model these
extensions in the Y-Bus formulation for simplicity.

\paragraph*{Line Charging}  

A line connecting buses $n$ and $m$ may have a predefined line charge
$\widetilde{Y}^c$. Steady state AC models typically
assume that a line charge is evenly distributed across the line and
hence it is resonable to assign equal portions of its charge to both
sides of the line. This is incorporated in the Y-Bus matrix as
follows:
\[
\widetilde{Y}_{nn}^{b'} = \widetilde{Y}_{nn}^b+\widetilde{Y}_{nm}^c/2,
\]
\[
\widetilde{Y}_{mm}^{b'} = \widetilde{Y}_{mm}^b+\widetilde{Y}_{nm}^c/2.
\]

\paragraph*{Transformers}  

A transformer connecting bus $n$ to bus $m$ can be modeled as a line
with modifications to the Y-Bus matrix.  The properties of the
transformer are captured by a complex number $\widetilde{T}_{nm} =
|\widetilde{T}| \angle s^\circ$, where $|\widetilde{T}|$ is the tap
ratio from $n$ to $m$ and $s^\circ$ is the phase shift.  It is worth
noting that the direction of a transformer-line is very important to
model the tap ratio and phase shift properly.  A transformer is
modeled in the Y-Bus matrix as follows:
\[
\widetilde{Y}_{nn}^{b'} = \widetilde{Y}_{nn}^b - \widetilde{Y}_{nm} +
\widetilde{Y}_{nm}/|\widetilde{T}_{nm}|^2,
\]
\[
\widetilde{Y}_{nm}^{b'} = \widetilde{Y}_{nm}/\widetilde{T}_{nm}^*,
\]
\[
\widetilde{Y}_{mn}^{b'} = \widetilde{Y}_{mn}/\widetilde{T}_{nm}.
\]
If a line charge exists, it must be applied before
the transformer calculation, i.e.,
\[
\widetilde{Y}_{nn}^{b'} = \widetilde{Y}_{nn}^b - \widetilde{Y}_{nm} +
(\widetilde{Y}_{nm}+\widetilde{Y}^c_{nm}/2)/|\widetilde{T}_{nm}|^2
\]

\paragraph*{Bus Shunts}  

A bus $n$ may have a shunt element which is modeled as a fixed
admittance to ground with a value of $\widetilde{Y}^{s}$. In the Y-Bus matrix, we have
\[
\widetilde{Y}_{nn}^{b'} = \widetilde{Y}_{nn}^b+\widetilde{Y}^{s}_n
\]
Unlike line charging, this extension is not affected by transformers,
since it applies to a bus and not a line.

\subsection{The Linearized DC Power Flow}

Many variants of the Linearized DC (LDC) model exist
\cite{nla.cat-vn193659,9780471586999,9780071447799,9780849373657}.  A
comprehensive review of all these variants is outside the scope of
this work but an in-depth discussion can be found in
\cite{Stott:2009bb}.  For brevity, we only review the simplest and
most popular variant of the LDC, which is derived from the AC
equations through a series of approximations justified by operational
considerations under normal operating conditions. In particular, the
LDC assumes that (1) the susceptance is large relative to the
conductance $|g| \ll |b|$; (2) the phase angle difference is small
enough to ensure $\sin(\theta^\circ_{n}-\theta^\circ_{m}) \approx
\theta^\circ_{n}-\theta^\circ_{m}$; and (3) the voltage magnitudes
$|\widetilde{V}|$ are close to $1.0$ and do not vary
significantly. Under these assumptions, Equations
(\ref{eq:acActiveLine}) and (\ref{eq:acReactiveLine}) reduce to
\begin{eqnarray}
p_{nm} = -b_{nm}(\theta^\circ_{n} - \theta^\circ_{m})
\end{eqnarray}
This simple linear formulation has been used in many frameworks for
decision support in power systems
\cite{Overbye:2004vb,4492805,Bienstock2007115,4762170,SSP1,PRVRP1}.
This traditional model is used as the baseline in the experimental
results.


\section{Linear-Programming Approximations}
\label{section:LAC}

This section presents linear-programming approximations of the AC
power flow equations. To understand the approximations, it is
important to distinguish between {\em hot-start} and {\em cold-start}
contexts \cite{Stott:2009bb}. In hot-start contexts, a solved AC
base-point solution is available and hence the model has at its
disposal additional information such as voltage magnitudes. In cold
start contexts, no such solved AC base-point solution is available and
it can be "maddeningly difficult" \cite{Overbye:2004vb} to obtain one
by simulation of the network. Hot-start models are well-suited for
applications in which the network topology is relatively stable, e.g.,
in LMP-base market calculations, optimal line switching, distribution
configuration, and real-time security constrained economic
dispatch. Cold-start models are used when no operational network is
available, e.g., in long-term planning studies. We also introduce the
concept of {\em warm-start} contexts, in which the model has at its
disposal target voltages (e.g., from normal operating conditions) but
an actual solution may not exist for these targets. Warm-start models
are particularly useful for power restoration applications in which
the goal is to return to normal operating conditions as quickly as
possible.  This section presents the hot-start, warm-start, and cold-start
models in stepwise refinements. It also discusses how models can be
generalized to include generation and load shedding, remove the slack
bus, impose constraints on voltages and reactive power, and capacity
constraints on the lines, all which are fundamental for many
applications. 

\subsection{AC Power Flow Behavior}

Before presenting the models, it is useful to review the behavior of
AC power flows, which is the main driver in the derivation. The
high-level behavior of power systems is often characterized by two
rules of thumb in the literature: (1) phase angles are the primary
factor in determining the flow active power; (2) differences in
voltage magnitudes are the primary factor in determining the flow of
reactive power \cite{9780070612938}. We examine these properties
experimentally.

The experiments make two basic assumptions: (1) In the per unit
system, voltages do not vary far from a magnitude of 1.0 and angle of 0.0; 
(2) The magnitude of a line conductance is much
smaller than the magnitude of the susceptance, i.e., $|g| \ll |b|$. We
can then explore the bounds of the power flow equations
(\ref{eq:acActiveLine}) and (\ref{eq:acReactiveLine}), when the
voltages are in the following bounds: $|\widetilde{V}_n| = 1.0,
|\widetilde{V}_m| \in (1.2,0.8), \theta^\circ_n - \theta^\circ_m \in
(-\pi/6,\pi/6)$.  These bounds are intentionally generous so 
that the power flow behavior within and outside normal 
operating conditions may be illustrated.

\begin{figure}[t]
\center
    \includegraphics[width=4.5cm]{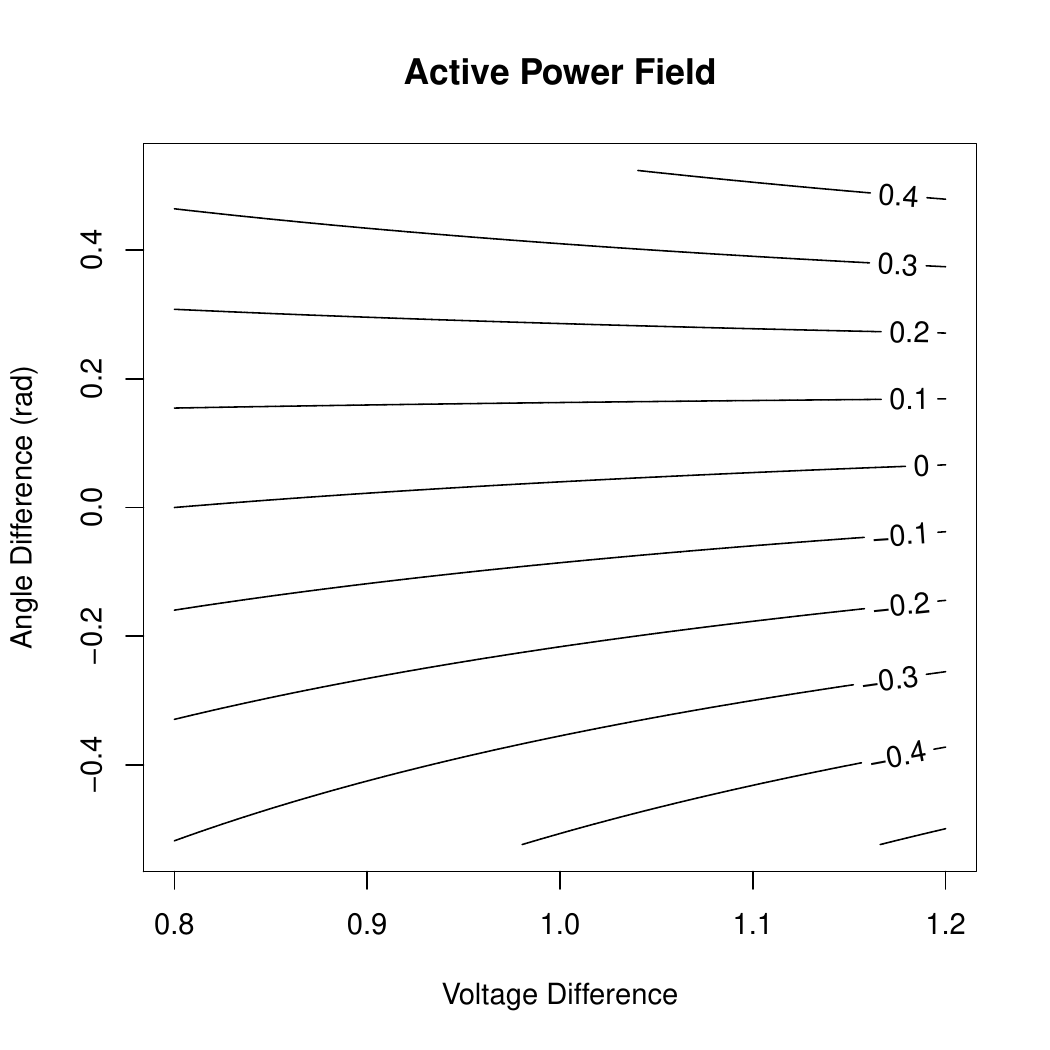}
    \hspace{-0.4cm}
    \includegraphics[width=4.5cm]{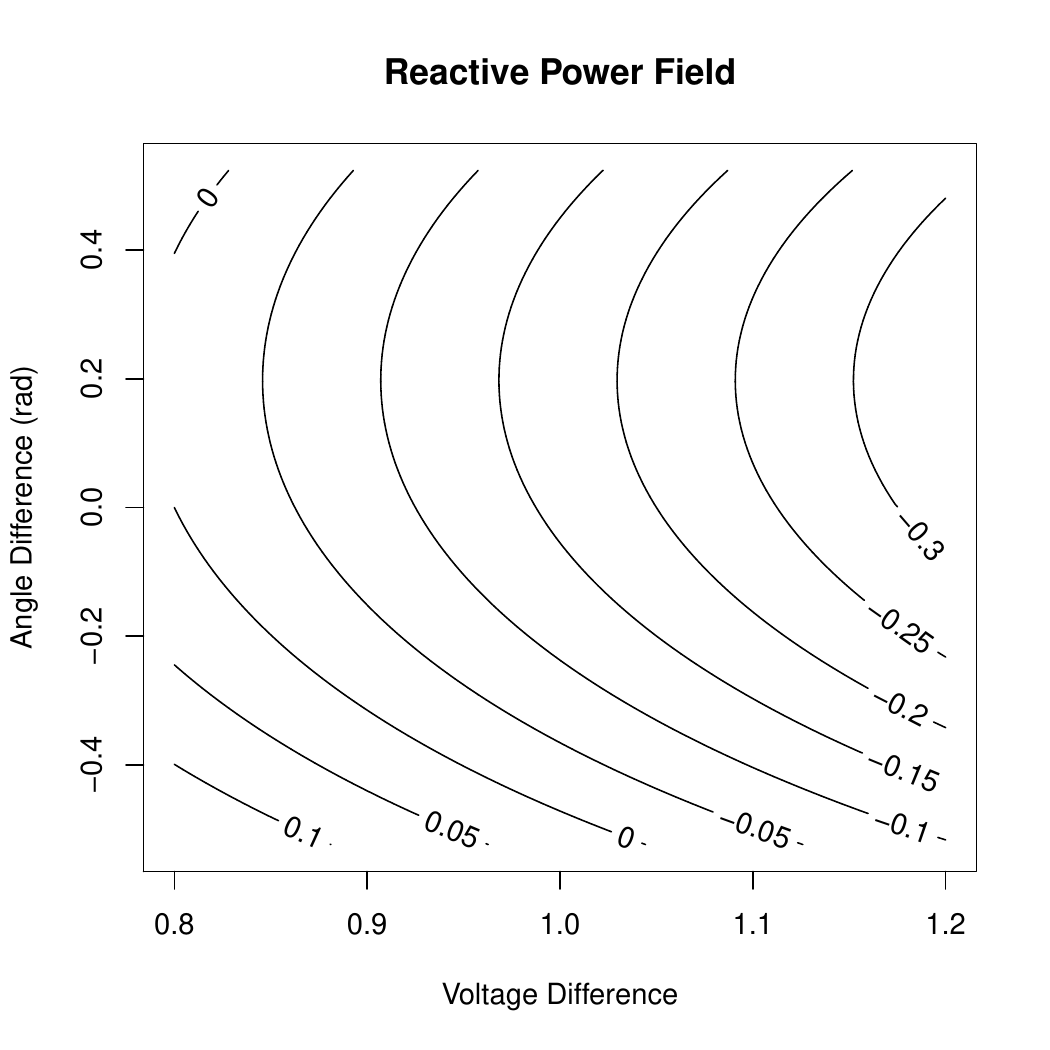} 
    \vspace{-0.6cm}
    \caption{Power Flow Contour of Active (left) and Reactive (right)
      Power with $g=0.2$ and $b=-1$.}
    \vspace{-0.4cm}
\label{fig:acPowerFields}
\end{figure}

Figure \ref{fig:acPowerFields} presents the contour of the active
power (left) and reactive power (right) equations for a line $\langle n,m \rangle$
under these assumptions when $\widetilde{Y}_{nm} = 0.2 -i1$.  The
contour lines indicate significant changes in power flow.  Consider
first the active power plot (left). For a fixed voltage, varying the
phase angle difference induces significant changes in active power as
many lines are crossed. In contrast, for a fixed phase angle
difference, varying the voltage has limited impact on the active
power, since few lines are crossed.  Hence, the plot indicates that
phase angle differences are the primary factor of active power flow
while voltage differences have only a small effect.  The situation is
quite different for reactive power (right plot).  For a fixed voltage,
varying the phase angle difference induces some significant changes in
reactive power as around four lines can be crossed. But, if the phase
angle difference is fixed, varying voltage induces even more
significant changes in reactive power since as many as seven lines may
now be crossed.  Hence, changes in voltages are the primary factor of
reactive power flows but the phase angle differences also have a
significant influence.

\subsection{The Hot-Start LPAC Model}

The linear-programming approximation of the AC Power flow
equations in a hot-start context is based on three ideas:
\begin{enumerate}
\item It uses the voltage magnitude $|\widetilde{V}^h_n|$ from the AC
  base-point solution at bus $n$;
\item It approximates $\sin(x)$ by $x$;
\item It uses a convex approximation of the cosine.\footnote{The domain of the cosine should not exceed 
the range $(-\pi/2,\pi/2)$ to ensure convexity. This range is generous for AC power flows.} 
\end{enumerate}
Let $\widehat{\cos}(\theta^\circ_{n} - \theta^\circ_{m})$ denote the
convexification of $\cos(\theta^\circ_{n} -
\theta^\circ_{m})$ in the range $(-\pi/2,\pi/2)$, then the linear-programming approximation solves
the line flow constraints
\begin{eqnarray}
\hat{p}^h_{nm} &=& |\widetilde{V}^h_n|^{2}g_{nm} - |\widetilde{V}^h_n||\widetilde{V}^h_m|g_{nm}\widehat{\cos}(\theta^\circ_{n} - \theta^\circ_{m}) \nonumber \\ && 
-|\widetilde{V}^h_n||\widetilde{V}^h_m|b_{nm}(\theta^\circ_{n} - \theta^\circ_{m}) \label{eq:acActiveLinePWL} \\
\hat{q}^h_{nm} &=& -|\widetilde{V}^h_n|^{2}b_{nm} + |\widetilde{V}^h_n||\widetilde{V}^h_m|b_{nm}\widehat{\cos}(\theta^\circ_{n} - \theta^\circ_{m}) \nonumber \\ && 
- |\widetilde{V}^h_n||\widetilde{V}^h_m|g_{nm}(\theta^\circ_{n} - \theta^\circ_{m}) \label{eq:acReactiveLinePWL}
\end{eqnarray}
\noindent
The details of the convexification are given in Appendix
\ref{section:lpcos}.  The hot-start model, presented in Model \ref{model:LPAChot}, is a linear program that
replaces the AC power equations by Equations \ref{eq:acActiveLinePWL}
and \ref{eq:acReactiveLinePWL} and thus captures an approximation of
reactive power in a linear formulation.  

A complete linear program for this formulation is presented in Model
\ref{model:LPAChot}.  The inputs to the model are: (1) A power
network ${\cal PN} = \langle N, L, G, s \rangle$, where $N$ is the set
of buses, $L$ is the set of lines, $G$ is the set of
voltage-controlled generators, $s$ is the slack bus; (2) the
voltage magnitudes $|V^h|$ for the buses and (3) the number of segments $cs$ for
approximating the cosine function.  The objective
(M\ref{model:LPAChot}.1) maximizes the cosine approximation to make
it as close as possible to the true cosine value.  Constraints
(M\ref{model:LPAChot}.2) model the slack bus, which has a fixed phase angle.  
Constraints (M\ref{model:LPAChot}.3)
and (M\ref{model:LPAChot}.4) model KCL on the buses.  
Like in AC power flow models, the KCL constraints are not enforced on the slack bus 
for both active and reactive power and on voltage-controlled generators for reactive
power.  Constraints (M\ref{model:LPAChot}.5) and
(M\ref{model:LPAChot}.6) capture the approximate line flows from
Equations (\ref{eq:acActiveLinePWL}) and (\ref{eq:acReactiveLinePWL}).
Finally, Constraints (M\ref{model:LPAChot}.7) define a system of
inequalities capturing the piecewise-linear approximation of the
cosine terms in the domain $(-\pi/3, \pi/3)$ using $cs$ line segments
for each line in the power network.

To our knowledge, this hot-start model 
is the first linear formulation that captures the cosine contribution to 
reactive power.  However, fixing the voltage
magnitudes, $|\widetilde{V}^h|$, in the power flow equations may be
too restrictive in many applications.  In the remaining sections, we
remove this restriction.

\begin{model}[t]
\footnotesize
\begin{tabbing}
123\= ${\cal PN} = \langle N, L, G, s \rangle$1 \= - phase angle on bus $i$ (radians)\=\kill
{\bf Inputs:}\\	
\> ${\cal PN} = \langle N, L, G, s \rangle$ \> - the power network \\
\> $|\widetilde{V}^h|$ \> - voltage magnitudes from a base-point solution \\
\> $cs$    \> -  cosine approximation segment count\\
{\bf Variables:}\\
\> $\theta^\circ_{n} \in (-\infty, \infty)$ \> - phase angle on bus $n$ (radians)\\
\> $\widehat{cos}_{nm} \in (0, 1)$ \> - Approximation of $\cos(\theta^\circ_{n} - \theta^\circ_{m})$\\
{\bf Maximize:}\\
123\= $\hat{q}^h_{nm} \!\! = \! -|\widetilde{V}^h_n|^{2}b_{nm} \!-\! |\widetilde{V}^h_n||\widetilde{V}^h_m|(g_{nm}(\theta^\circ_{n} - \theta^\circ_{m}) -  b_{nm}\widehat{\cos}_{nm})$ \=\=(2)\kill
\> $\displaystyle \sum_{\langle n,m \rangle \in L} \widehat{cos}_{nm}$ \>\> (M\ref{model:LPAChot}.1) \\
g{\bf Subject to:}\\
\> $\theta^\circ_{s} = 0$ \>\> (M\ref{model:LPAChot}.2)\\
\> $p_{n} = \displaystyle \sum_{m \in N}^{n \neq m} \hat{p}^h_{nm}  \;\; \forall n \in N \;\; n \neq s$ \>\>(M\ref{model:LPAChot}.3)\\ 
\> $q_{n} = \displaystyle \sum_{m \in N}^{n \neq m} \hat{q}^h_{nm}  \;\; \forall n \in N\ \;\; n \neq s \;\; n \not\in G$ \>\>(M\ref{model:LPAChot}.4)\\ 
\> $\forall \langle n,m \rangle,\langle m,n \rangle \in L$ \\
\> $\hat{p}^h_{nm} \!\! = \! |\widetilde{V}^h_n|^{2}g_{nm} \!-\! |\widetilde{V}^h_n||\widetilde{V}^h_m|(g_{nm}\widehat{\cos}_{nm} + b_{nm}(\theta^\circ_{n} - \theta^\circ_{m}))$
\>\>(M\ref{model:LPAChot}.5)\\ 
\> $\hat{q}^h_{nm} \!\! = \! -|\widetilde{V}^h_n|^{2}b_{nm} \!-\! |\widetilde{V}^h_n||\widetilde{V}^h_m|(g_{nm}(\theta^\circ_{n} - \theta^\circ_{m}) -  b_{nm}\widehat{\cos}_{nm})$
\>\>(M\ref{model:LPAChot}.6)\\
\> PWL$\langle COS \rangle$$(\widehat{\cos}_{nm}, (\theta^\circ_{n} - \theta^\circ_{m}), -\pi/3, \pi/3, cs)$
\>\>(M\ref{model:LPAChot}.7) 
\end{tabbing}
\caption{The Hot-Start LPAC Model.}
\label{model:LPAChot}
\end{model}

\subsection{The Warm-Start LPAC Model}

This section derives the warm-start LPAC model, i.e., a
Linear-Programming model of the AC power flow equations for the
warm-start context.  The warm-start context assumes that some target
voltages $|\widetilde{V}^t|$ are available for all buses except
voltage-controlled generators whose voltage magnitudes
$|\widetilde{V}^g|$ are known. The network {\bf must} operate close
(e.g., $\pm 0.1$ Volts p.u.)  to these target voltages, since
otherwise the hardware may be damaged or voltages may collapse.

The warm-start LPAC model is based on two key ideas:
\begin{enumerate}
\item The active power approximation is the same as in the hot-start
  model, with the target voltages replacing the voltages in the
  base-point solution;

\item The reactive power approximation reasons about voltage
  magnitudes, since changes in voltages are the primary factor of
  reactive power flows.
\end{enumerate}
To derive the reactive power approximation in the warm-start LPAC
model, let $\phi$ be the difference between the target voltage and the
true value, i.e.,
\[
|\widetilde{V}| = |\widetilde{V}^t| + \phi.
\]
Substituting in Equation \ref{eq:acReactiveLine}, we obtain
\begin{eqnarray}
q_{nm} &=& - (|\widetilde{V}^t_n|^{2} + 2|\widetilde{V}^t_n|\phi_n +  \phi_n^2)b_{nm} - \nonumber \\
              &  & (|\widetilde{V}^t_n||\widetilde{V}^t_m| + |\widetilde{V}^t_n|\phi_m + |\widetilde{V}^t_m|\phi_n + \phi_n\phi_m) \nonumber \\
              &  & (g_{nm}\sin(\theta^\circ_{n} - \theta^\circ_{m}) -  b_{nm}\cos(\theta^\circ_{n} - \theta^\circ_{m}))
\end{eqnarray}

\noindent
We can divide this expression into two parts 
\begin{eqnarray}
q_{nm} &=& q^t_{nm} + q^{\Delta}_{nm} \label{eq:acReactiveLineDeltaForm}
\end{eqnarray}
where $q^t_{nm}$ is Equation \ref{eq:acReactiveLine} with
$|\widetilde{V}| = |\widetilde{V}^t|$ and $q^{\Delta}_{nm}$ captures
the remaining terms, i.e.,
\begin{eqnarray}
q^{\Delta}_{nm} &=&  - (2|\widetilde{V}^t_n|\phi_n +  \phi_n^2)b_{nm} - \nonumber \\
              &&(|\widetilde{V}^t_n|\phi_m + |\widetilde{V}^t_m|\phi_n + \phi_n\phi_m) \nonumber \\
              &&(g_{nm}\sin(\theta^\circ_{n} - \theta^\circ_{m}) -  b_{nm}\cos(\theta^\circ_{n} - \theta^\circ_{m}))
\end{eqnarray}
Equation \ref{eq:acReactiveLineDeltaForm} is equivalent to Equation
\ref{eq:acReactiveLine} and must be linearized to obtain the LPAC
model.

The $q^t_{nm}$ part has target voltages and may thus be approximated
like $\hat{q}^h_{nm}$. The $q^{\Delta}_{nm}$ is more challenging as it
contains nonlinear and non-convex terms such as
$\phi_n\phi_m\cos(\theta^\circ_{n}-\theta^\circ_{m})$. We approximate
$q^{\Delta}_{nm}$ using the linear terms of the Taylor series of $q^{\Delta}_{nm}$
at $\phi_n=0, \phi_m=0, \theta^\circ_{n} - \theta^\circ_{m} = 0$ to obtain
\begin{eqnarray}
\hat{q}^{\Delta}_{nm} &=& -(2|\widetilde{V}^t_n|\phi_n)b_{nm} + (|\widetilde{V}^t_n|\phi_m + |\widetilde{V}^t_m|\phi_n)b_{nm} 
\end{eqnarray}
or, equivalently,
\begin{eqnarray}
\hat{q}^{\Delta}_{nm} &=& -|\widetilde{V}^t_n| b_{nm} (\phi_n - \phi_m) - (|\widetilde{V}^t_n| - |\widetilde{V}^t_m|) b_{nm} \phi_n \label{eq:acReactiveLineWarmApprox}
\end{eqnarray}
A complete linear program for this formulation is presented in Model
\ref{model:LPACwarm}.  The inputs to the model are similar to Model \ref{model:LPAChot}, with hot start voltages $|\widetilde{V}^h|$ replaced by  target
voltages $|\widetilde{V}^t|$.  The objective
(M\ref{model:LPACwarm}.1) maximizes the cosine approximation to make
it as close as possible to the true cosine value.  Constraints
(M\ref{model:LPACwarm}.2) model the slack bus, which has a fixed voltage
and phase angle.  Constraints (M\ref{model:LPACwarm}.3) capture the
voltage-controlled generators which, by definition, do not vary from
their voltage target $|V^t|$.  Constraints (M\ref{model:LPACwarm}.4)
and (M\ref{model:LPACwarm}.5) model KCL on the buses, as well as the
effects of voltage change presented in Equation
(\ref{eq:acReactiveLineWarmApprox}).  Like in AC power flow models,
the KCL constraints are not enforced on the slack bus for both active
and reactive power and on voltage-controlled generators for reactive
power.  Constraints (M\ref{model:LPACwarm}.6) and
(M\ref{model:LPACwarm}.7) capture the approximate line flows from
Equations (\ref{eq:acActiveLinePWL}) and (\ref{eq:acReactiveLinePWL}).
Constraints (M\ref{model:LPACwarm}.8) model the effects of voltage
change presented in Equation (\ref{eq:acReactiveLineWarmApprox}).
Finally, Constraints (M\ref{model:LPACwarm}.9) define a system of
inequalities capturing the piecewise-linear approximation of the
cosine terms in the domain $(-\pi/3, \pi/3)$ using $cs$ line segments
for each line in the power network.

\begin{model}[t]
\footnotesize
\begin{tabbing}
123\= ${\cal PN} = \langle N, L, G, s \rangle$1 \= - phase angle on bus $i$ (radians)\=\kill
{\bf Inputs:}\\	
\> ${\cal PN} = \langle N, L, G, s \rangle$ \> - the power network \\
\> $|\widetilde{V}^t|$ \> - target voltage magnitudes \\
\> $cs$    \> -  cosine approximation segment count\\
{\bf Variables:}\\
\> $\theta^\circ_{n} \in (-\infty, \infty)$ \> - phase angle on bus $n$ (radians)\\
\> $\phi_{n} \in (-|V^t|, \infty)$ \> - voltage change on bus $n$ (Volts p.u.)\\
\> $\widehat{cos}_{nm} \in (0, 1)$ \> - Approximation of $\cos(\theta^\circ_{n} - \theta^\circ_{m})$\\
{\bf Maximize:}\\
123\= $\hat{q}^t_{nm} \!\! = \! -|\widetilde{V}^t_n|^{2}b_{nm} \!-\! |\widetilde{V}^t_n||\widetilde{V}^t_m|(g_{nm}(\theta^\circ_{n} - \theta^\circ_{m}) -  b_{nm}\widehat{\cos}_{nm})$ \=\=(2)\kill
\> $\displaystyle \sum_{\langle n,m \rangle \in L} \widehat{cos}_{nm}$ \>\> (M\ref{model:LPACwarm}.1) \\
{\bf Subject to:}\\
\> $\theta^\circ_{s} = 0, \phi_{s} = 0$ \>\> (M\ref{model:LPACwarm}.2)\\
\> $\phi_{i} = 0 \;\; \forall i \in G$ \>\> (M\ref{model:LPACwarm}.3) \\
\> $p_{n} = \displaystyle \sum_{m \in N}^{n \neq m} \hat{p}^t_{nm}  \;\; \forall n \in N \;\; n \neq s$ \>\>(M\ref{model:LPACwarm}.4)\\ 
\> $q_{n} = \displaystyle \sum_{m \in N}^{n \neq m} \hat{q}^t_{nm} + \hat{q}^{\Delta}_{nm} \;\; \forall n \in N\ \;\; n \neq s \;\; n \not\in G$ \>\>(M\ref{model:LPACwarm}.5)\\ 
\> $\forall \langle n,m \rangle,\langle m,n \rangle \in L$ \\
\> $\hat{p}^t_{nm} \!\! = \! |\widetilde{V}^t_n|^{2}g_{nm} \!-\! |\widetilde{V}^t_n||\widetilde{V}^t_m|(g_{nm}\widehat{\cos}_{nm} + b_{nm}(\theta^\circ_{n} - \theta^\circ_{m}))$
\>\>(M\ref{model:LPACwarm}.6)\\ 
\> $\hat{q}^t_{nm} \!\! = \! -|\widetilde{V}^t_n|^{2}b_{nm} \!-\! |\widetilde{V}^t_n||\widetilde{V}^t_m|(g_{nm}(\theta^\circ_{n} - \theta^\circ_{m}) -  b_{nm}\widehat{\cos}_{nm})$
\>\>(M\ref{model:LPACwarm}.7)\\
\> $\hat{q}^{\Delta}_{nm} = -|\widetilde{V}^t_n| b_{nm} (\phi_n - \phi_m) - (|\widetilde{V}^t_n| - |\widetilde{V}^t_m|) b_{nm} \phi_n$
\>\>(M\ref{model:LPACwarm}.8)  \\
\> PWL$\langle COS \rangle$$(\widehat{\cos}_{nm}, (\theta^\circ_{n} - \theta^\circ_{m}), -\pi/3, \pi/3, cs)$
\>\>(M\ref{model:LPACwarm}.9) 
\end{tabbing}
\caption{The Warm-Start LPAC Model.}
\label{model:LPACwarm}
\end{model}

\subsection{The Cold-Start LPAC Model}

We now conclude by presenting the cold-start LPAC model. In a
cold-start context, no target voltages are available and voltage
magnitudes are approximated by 1.0, except for voltage-controlled
generators whose voltages are given by $|\widetilde{V}_n^g|$ $(n \in
G)$. The cold-start LPAC model is then derived from the warm-start
LPAC model by fixing $|\widetilde{V}^t_i|=1$ for all $i \in N$. 
Equation (\ref{eq:acReactiveLineWarmApprox}) then reduces to
\begin{eqnarray}
\hat{q}^{\Delta}_{nm} &=& - b_{nm}(\phi_n - \phi_m)   \label{eq:acReactiveLineColdApproxFinal}
\end{eqnarray}
Figure \ref{model:LPACcold} presents the cold-start LPAC model, which
is very close to the warm-start model. Note that Constraints
(M\ref{model:LPACcold}.3) use $\phi_i$ to fix the voltage magnitudes
of generators.

\begin{model}[t]
\footnotesize
\begin{tabbing}
123\= ${\cal PN} = \langle N, L, G, s \rangle$1 \= - phase angle on bus $i$ (radians)\=\kill
{\bf Inputs:}\\	
\> ${\cal PN} = \langle N, L, G, s \rangle$ \> - the power network \\
\> $cs$    \> -  cosine approximation segment count\\
{\bf Variables:}\\
\> $\theta^\circ_{n} \in (-\infty, \infty)$ \> - phase angle on bus $n$ (radians)\\
\> $\phi_{n} \in (-|V^t|, \infty)$ \> - voltage change on bus $n$ (Volts p.u.)\\
\> $\widehat{cos}_{nm} \in (0, 1)$ \> - Approximation of $\cos(\theta^\circ_{n} - \theta^\circ_{m})$\\
{\bf Maximize:}\\
123\= $\hat{q}^t_{nm} \!\! = \! -|\widetilde{V}^t_n|^{2}b_{nm} \!-\! |\widetilde{V}^t_n||\widetilde{V}^t_m|(g_{nm}(\theta^\circ_{n} - \theta^\circ_{m}) -  b_{nm}\widehat{\cos}_{nm})$ \=\=(2)\kill
\> $\displaystyle \sum_{\langle n,m \rangle \in L} \widehat{cos}_{nm}$ \>\> (M\ref{model:LPACcold}.1) \\
{\bf Subject to:}\\
\> $\theta^\circ_{s} = 0, \phi_{s} =  |\widetilde{V}^g_s| - 1.0$ \>\> (M\ref{model:LPACcold}.2)\\
\> $\phi_{i} = |\widetilde{V}^g_i| - 1.0 \;\; \forall i \in G$ \>\> (M\ref{model:LPACcold}.3) \\
\> $p_{n} = \displaystyle \sum_{m \in N}^{n \neq m} \hat{p}^t_{nm}  \;\; \forall n \in N \;\; n \neq s$ \>\>(M\ref{model:LPACcold}.4)\\ 
\> $q_{n} = \displaystyle \sum_{m \in N}^{n \neq m} \hat{q}^t_{nm} + \hat{q}^{\Delta}_{nm} \;\; \forall n \in N\ \;\; n \neq s \;\; n \not\in G$ \>\>(M\ref{model:LPACcold}.5)\\ 
\> $\forall \langle n,m \rangle,\langle m,n \rangle \in L$ \\
\> $\hat{p}^t_{nm} \!\! = \! g_{nm} \!-\! g_{nm}\widehat{\cos}_{nm} - b_{nm}(\theta^\circ_{n} - \theta^\circ_{m})$
\>\>(M\ref{model:LPACcold}.6)\\ 
\> $\hat{q}^t_{nm} \!\! = \! -b_{nm} \!-\! g_{nm}(\theta^\circ_{n} - \theta^\circ_{m}) + b_{nm}\widehat{\cos}_{nm}$
\>\>(M\ref{model:LPACcold}.7)\\
\> $\hat{q}^{\Delta}_{nm} = -b_{nm} (\phi_n - \phi_m)$ 
\>\>(M\ref{model:LPACcold}.8) \\
\> PWL$\langle COS \rangle$$(\widehat{\cos}_{nm}, (\theta^\circ_{n} - \theta^\circ_{m}), -\pi/3, \pi/3, cs)$
\>\>(M\ref{model:LPACcold}.9) 
\end{tabbing}
\caption{The Cold-Start LPAC Model.}
\label{model:LPACcold}
\end{model}

\subsection{Extensions to the LPAC Model}

The LPAC models can be used to solve the AC power flow equations
approximately in a variety of contexts. This section reviews how to
generalize the LPAC models for applications in disaster management,
reactive voltage support, transmission planning, and vulnerability
analysis. The extensions are illustrated on the warm-start model but
can be similarly applied to the cold-start model.

\paragraph*{Generators} The LPAC model can easily be generalized to
include ranges for generators: Simply remove the generator from $G$
and place operating limits on the $p$ and $q$ variables for that
bus. In this formulation, voltage-controlled generators can also be
accommodated by fixing $\phi_n$ to zero at bus $n$.

\paragraph*{Removing the Slack Bus} By necessity, AC solvers use a
slack bus to ensure the flow balance in the network when the total
power consumption is not known a priori (e.g., due to line losses). As
a consequence, the LPAC model depicted in Figure \ref{model:LPACwarm}
also uses a slack bus so that the AC and LPAC models can be accurately
compared in our experimental results. However, it is important to
emphasize that the LPAC model does not need a slack bus and the only
reason to include a slack bus in this model is to allow for meaningful
comparisons between the LPAC and AC models. As discussed above, the
LPAC model can easily include a range for each generator, thus
removing the need for a slack bus.

\paragraph*{Load Shedding} For applications in power restoration
(e.g., \cite{SSP1,PRVRP1,PES2}), the LPAC model can also
integrate load shedding: Simply transform the loads into decision
variables with an upper bound and maximize the load served. The cosine
maximization should also be included in the objective but with a
smaller weight. Section \ref{section-results-restoration} reports
experimental results on such a power restoration model.

\paragraph*{Modeling Additional Constraints} In practice, feasibility
constraints may exist on the acceptable voltage range, the reactive
injection of a generator, or line flow capacities.  Because Model
\ref{model:LPACwarm} is a linear program, it can incorporate such
constraints. For instance, constraint
\[
\underline{|V|} \leq |V_n^t| + \phi_n  \;\; \forall n \in N
\]
ensures that voltages are above a certain limit $\underline{|V|}$,
constraint 
\[
\displaystyle \sum_{m \in N}^{n \neq m} \hat{q}^t_{nm}
+\hat{q}^\Delta_{nm} \leq \overline{q_n} \;\; \forall n \in G
\]
limits the maximum reactive injection bounds at bus $n$ to
$\overline{q_n}$. Finally, let $\overline{|{S}_{nm}|}$ be
the maximum apparent power on a line from bus $n$ to bus $m$.  Then,
constraint
\[
(\hat{p}^t_{nm})^2 + (\hat{q}^t_{nm} +\hat{q}^\Delta_{nm})^2 \leq \overline{|{S}_{nm}|}^2
\]
ensures that line flows are feasible in the LPAC model. The quadratic
functions can be approximated by piecewise-linear constrains
(e.g., \cite{PES1}). 

\section{Accuracy of the LPAC Model}
\label{section:Accuracy}

This section evaluates the accuracy of the LPAC models by comparing them to an ideal nonlinear AC power flow.\footnote{For consistency, the LPAC models are extended to include line charging, bus shunts, and transformers, as
discussed in Section \ref{Section:particalnetworks}.} It includes a
detailed analysis of the model accuracy (Section
\ref{section:exp:accuracy}) and an investigation of alternative
approximations (Section \ref{section:exp:approx}).  The experiments
were performed on nine traditional power-system benchmarks which come
from the IEEE test systems \cite{IEEEBench} and {\sc Matpower}
\cite{MurilloSanchez:1997ws}.  The AC power flow equations were solved
with a Newton-Raphson solver which was validated using {\sc Matpower}.
The LPAC models use 20 line segments in the cosine approximation and
all of the models solved in less than 1 second on a 2.5 GHz Intel
processor. The results also include a modified version of the IEEEdd17
benchmark, called IEEEdd17m.  The original IEEEdd17 has the slack bus
connected to the network by a transformer with
$|\widetilde{T}|=1.05$. The nonlinear behavior of transformers induces
some loss of accuracy in the LPAC model and, because this error occurs
at the slack bus in IEEEdd17, it affects all buses in the
network. IEEEdd17m resolves this issue by setting
$|\widetilde{T}|=1.00$ and the slack bus voltage to $1.05$.  As the results indicate, 
this equivalent formulation is significantly better for the LPAC model.

\subsection{Accuracy of The LPAC Models}
\label{section:exp:accuracy}

This section reports empirical evaluations of the LDC and LPAC models
in cold-start and warm-start contexts.  It reports aggregate
statistics for active power (Table \ref{tbl:ldcs_active}), bus phase
angles (Table \ref{tbl:ldcs_angle}), reactive power (Table
\ref{tbl:ldcs_reactive}), and voltage magnitudes (Table
\ref{tbl:ldcs_voltage}).  Data for the LDC model is necessarily
omitted from Tables \ref{tbl:ldcs_reactive} and \ref{tbl:ldcs_voltage}
as reactive power and voltages are not captured by that model.  In
each table, two aggregate values are presented: Correlation (corr) and
absolute error ($\Delta$).  The units of the absolute error are
presented in the headings. Both average ($\mu$) and worst-case
($\max$) values are presented. The worst case can often be misleading:
For example a very large value may actually be a very small relative
quantity. For this reason, the tables show the relative error
($\delta$) of the value selected by the $\max$ operator using the
$\arg$-$\max$ operator.  The relative error is a percentage and is
unit-less.

\begin{table}[t]
\center
\caption{Accuracy of the LPAC Model: Active Power Flows.}
\begin{tabular}{|c||c||c|c|c|}
\hline
Benchmark & \multicolumn{4}{|c|}{Active Power (MW)} \\
 & Corr & $\mu(\Delta)$ & $\max(\Delta)$ & $\delta(\arg$-$\max(\Delta))$ \\
\hline

\multicolumn{5}{|c|}{The LDC Model}\\

\hline
ieee14 & 0.9994 & 1.392 & 10.64 & 6.783 \\
\hline
mp24 & 0.9989 & 5.659 & 19.7 & 23.65 \\
\hline
ieee30 & 0.9993 & 1.046 & 13.1 & 7.562  \\
\hline
mp30 & 0.9993 & 0.2964 & 2.108 & 19.36  \\
\hline
mp39 & 0.9995 & 7.341 & 43.64 & 6.527  \\
\hline
ieee57 & 0.9989 & 1.494 & 8.216 & 8.055 \\
\hline
ieee118 & 0.9963 & 3.984 & 56.3 & 44.74  \\
\hline
ieeedd17 & 0.9972 & 4.933 & 201.3 & 13.84  \\
\hline
ieeedd17m & 0.9975 & 4.779 & 191.1 & 13.23  \\
\hline
mp300 & 0.9910 & 11.09 & 418.5 & 90.02  \\
\hline

\multicolumn{5}{|c|}{The LPAC-Cold Model}\\

\hline
ieee14 & 0.9989 & 1.636 & 5.787 & 13.13 \\
\hline
mp24 & 0.9999 & 1.884 & 6.159 & 2.933 \\
\hline
ieee30 & 0.9998 & 0.5475 & 2.213 & 2.523 \\
\hline
mp30 & 0.9995 & 0.2396 & 1.641 & 15.07 \\
\hline
mp39 & 1.0000 & 2.142 & 8.043 & 3.288 \\
\hline
ieee57 & 0.9995 & 0.9235 & 4.674 & 9.728 \\
\hline
ieee118 & 1.0000 & 0.622 & 3.708 & 2.038 \\
\hline
ieeedd17 & 0.9999 & 1.827 & 30.38 & 2.088 \\
\hline
ieeedd17m & 0.9999 & 1.475 & 20.21 & 1.399 \\
\hline
mp300 & 0.9998 & 2.455 & 18 & 8.675 \\
\hline

\multicolumn{5}{|c|}{The LPAC-Warm Model}\\

\hline
ieee14 & 1.0000 & 0.1689 & 1.588 & 1.012 \\
\hline
mp24 & 1.0000 & 0.6621 & 2.041 & 1.01 \\
\hline
ieee30 & 1.0000 & 0.1847 & 2.433 & 1.405 \\
\hline
mp30 & 0.9999 & 0.1052 & 0.705 & 6.474 \\
\hline
mp39 & 1.0000 & 1.557 & 11.58 & 1.731 \\
\hline
ieee57 & 1.0000 & 0.2229 & 2.013 & 1.973 \\
\hline
ieee118 & 0.9999 & 0.4386 & 7.376 & 5.862 \\
\hline
ieeedd17 & 1.0000 & 0.58 & 22.5 & 1.547 \\
\hline
ieeedd17m & 1.0000 & 0.5725 & 21.73 & 1.504 \\
\hline
mp300 & 0.9999 & 1.195 & 52.84 & 11.37 \\
\hline

\end{tabular}
\label{tbl:ldcs_active}
\end{table}

Table \ref{tbl:ldcs_active} indicates uniform improvements in active
power flows, especially in the largest benchmarks IEEE118, IEEEdd17,
and MP300.  Significant errors are not uncommon for the linearized DC model
on large benchmarks \cite{Stott:2009bb} and are primarily caused by a
lack of line losses.  Due to its asymmetrical power flow equations and
the cosine approximation, the LPAC model captures line losses.

\begin{table}[t]
\center
\caption{Accuracy of the LPAC Model: Phase Angles.}
\begin{tabular}{|c||c||c|c|c|}
\hline
Benchmark & \multicolumn{4}{|c|}{Phase Angle (rad)} \\
 & Corr & $\mu(\Delta)$ & $\max(\Delta)$ & $\delta(\arg$-$\max(\Delta))$ \\
\hline

\multicolumn{5}{|c|}{The LDC Model}\\

\hline
ieee14 & 0.9993 & 0.02487 & 0.04258 & 15.22  \\
\hline
mp24 & 0.9997 & 0.01334 & 0.02037 & 15.23  \\
\hline
ieee30 & 0.9981 & 0.02831 & 0.04733 & 16.45  \\
\hline
mp30 & 0.9800 & 0.005658 & 0.01607 & 30.27  \\
\hline
mp39 & 0.9951 & 0.0283 & 0.05813 & 85.56  \\
\hline
ieee57 & 0.9898 & 0.02244 & 0.05958 & 24.1  \\
\hline
ieee118 & 0.9904 & 0.03452 & 0.09026 & 88.41 \\
\hline
ieeedd17 & 0.9892 & 0.115 & 0.1395 & 16.09  \\
\hline
ieeedd17m & 0.9920 & 0.0461 & 0.06924 & 41.88  \\
\hline
mp300 & 0.9752 & 0.3103 & 0.4244 & 975.7  \\
\hline

\multicolumn{5}{|c|}{The LPAC-Cold Model}\\

\hline
ieee14 & 0.9971 & 0.004525 & 0.01241 & 5 \\
\hline
mp24 & 0.9999 & 0.003539 & 0.008947 & 6.922 \\
\hline
ieee30 & 0.9965 & 0.007268 & 0.02413 & 8.386 \\
\hline
mp30 & 0.9782 & 0.006236 & 0.01804 & 33.99 \\
\hline
mp39 & 0.9989 & 0.006268 & 0.02314 & 34.06 \\
\hline
ieee57 & 0.9894 & 0.0179 & 0.05467 & 22.11 \\
\hline
ieee118 & 0.9994 & 0.003225 & 0.01354 & 9.633 \\
\hline
ieeedd17 & 0.9981 & 0.03648 & 0.05165 & 5.958 \\
\hline
ieeedd17m & 0.9990 & 0.007207 & 0.02682 & 4.522 \\
\hline
mp300 & 0.9984 & 0.01458 & 0.08086 & 38.49 \\
\hline

\multicolumn{5}{|c|}{The LPAC-Warm Model}\\

\hline
ieee14 & 1.0000 & 0.001448 & 0.001829 & 0.6914 \\
\hline
mp24 & 1.0000 & 0.001337 & 0.002203 & 2.156 \\
\hline
ieee30 & 1.0000 & 0.002345 & 0.002819 & 0.9629 \\
\hline
mp30 & 0.9998 & 0.001298 & 0.001774 & 4.775 \\
\hline
mp39 & 0.9999 & 0.005315 & 0.006241 & 4.273 \\
\hline
ieee57 & 1.0000 & 0.002711 & 0.00357 & 1.776 \\
\hline
ieee118 & 0.9999 & 0.005958 & 0.008366 & 2.526 \\
\hline
ieeedd17 & 0.9999 & 0.01492 & 0.01719 & 2.419 \\
\hline
ieeedd17m & 0.9999 & 0.008443 & 0.01059 & 2.33 \\
\hline
mp300 & 0.9997 & 0.03842 & 0.04502 & 18.51 \\
\hline

\end{tabular}

\label{tbl:ldcs_angle}
\end{table}

Table \ref{tbl:ldcs_angle} presents the aggregate statistics on bus
phase angles.  These results show significant improvements in accuracy
especially on larger benchmarks. The correlations are somewhat lower
than active power, but phase angles are quite challenging from a
numerical accuracy standpoint.

\begin{table}[t]
\center
\caption{Accuracy of the LPAC Model: Reactive Power Flows.}
\begin{tabular}{|c||c||c|c|c|}
\hline
Benchmark & \multicolumn{4}{|c|}{Reactive Power (MVar)} \\
 & Corr & $\mu(\Delta)$ & $\max(\Delta)$ & $\delta(\arg$-$\max(\Delta))$ \\
\hline

\multicolumn{5}{|c|}{The LPAC-Cold Model}\\

\hline
ieee14 & 0.9948 & 0.7459 & 2.561 & 14.92  \\
\hline
mp24 & 0.9992 & 1.505 & 5.245 & 9.309  \\
\hline
ieee30 & 0.997 & 0.4962 & 1.902 & 23.36  \\
\hline
mp30 & 0.9991 & 0.3135 & 0.8925 & 3.886 \\
\hline
mp39 & 0.9973 & 3.898 & 15.15 & 18.25 \\
\hline
ieee57 & 0.9991 & 0.5316 & 2.98 & 3.973 \\
\hline
ieee118 & 0.9991 & 0.7676 & 6.248 & 8.561 \\
\hline
ieeedd17 & 0.9789 & 3.989 & 48.65 & 40.58 \\
\hline
ieeedd17m & 0.9927 & 2.415 & 34.18 & 12.36 \\
\hline
mp300 & 0.9981 & 3.85 & 62.32 & 17.98 \\
\hline

\multicolumn{5}{|c|}{The LPAC-Warm Model}\\

\hline
ieee14 & 0.9895 & 0.8689 & 3.167 & 43.89 \\
\hline
mp24 & 0.9992 & 1.505 & 5.245 & 9.309 \\
\hline
ieee30 & 0.9975 & 0.3455 & 1.607 & 7.62 \\
\hline
mp30 & 0.9991 & 0.3135 & 0.8925 & 3.886 \\
\hline
mp39 & 0.9971 & 4.03 & 15.75 & 18.98 \\
\hline
ieee57 & 0.9995 & 0.3853 & 1.46 & 5.67 \\
\hline
ieee118 & 0.9992 & 0.6326 & 6.109 & 6.808  \\
\hline
ieeedd17 & 0.9791 & 3.985 & 48.37 & 40.34 \\
\hline
ieeedd17m & 0.9927 & 2.409 & 33.96 & 12.28 \\
\hline
mp300 & 0.9943 & 3.584 & 162 & 45.05 \\
\hline

\end{tabular}
\label{tbl:ldcs_reactive}
\end{table}

\begin{figure}[t]
\center
    \includegraphics[width=4.5cm]{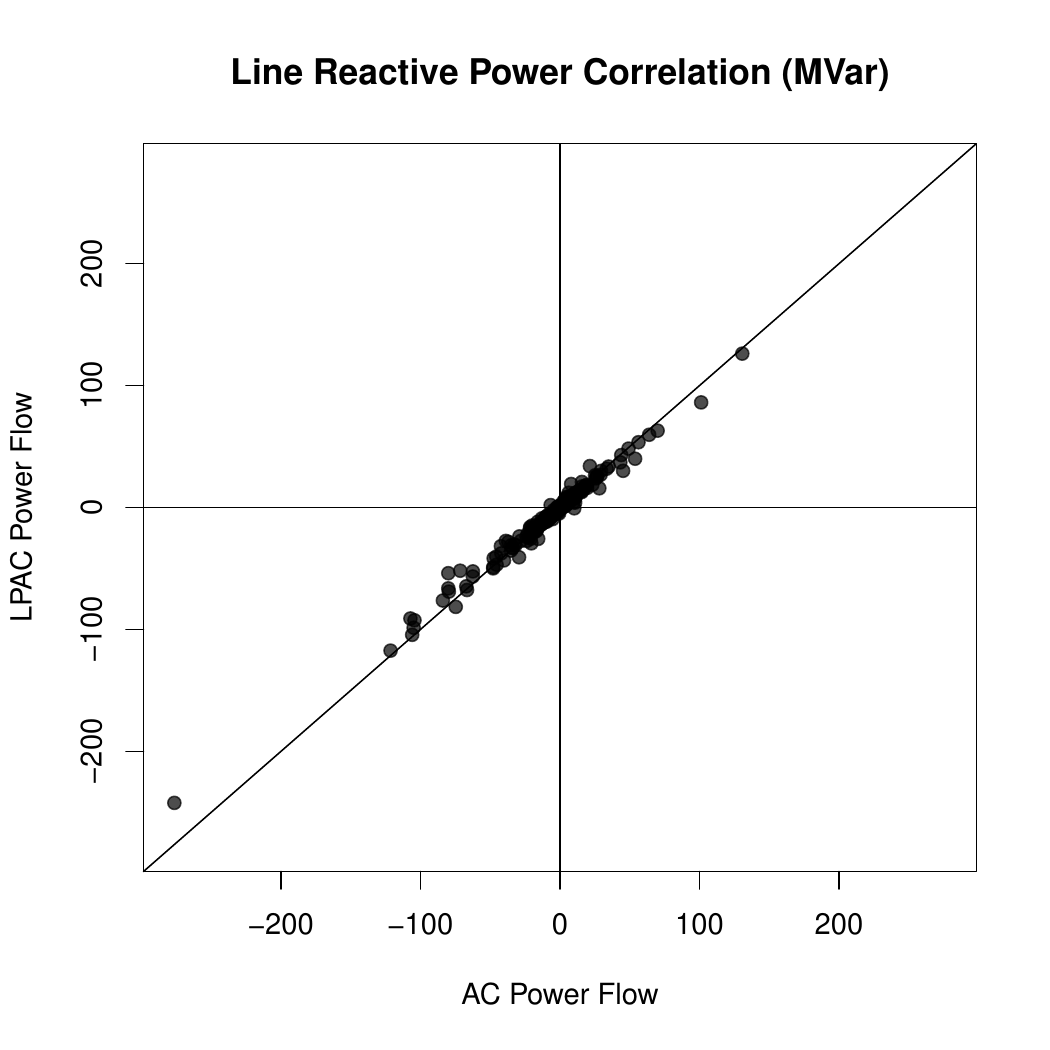}  
    \hspace{-0.4cm}
    \includegraphics[width=4.5cm]{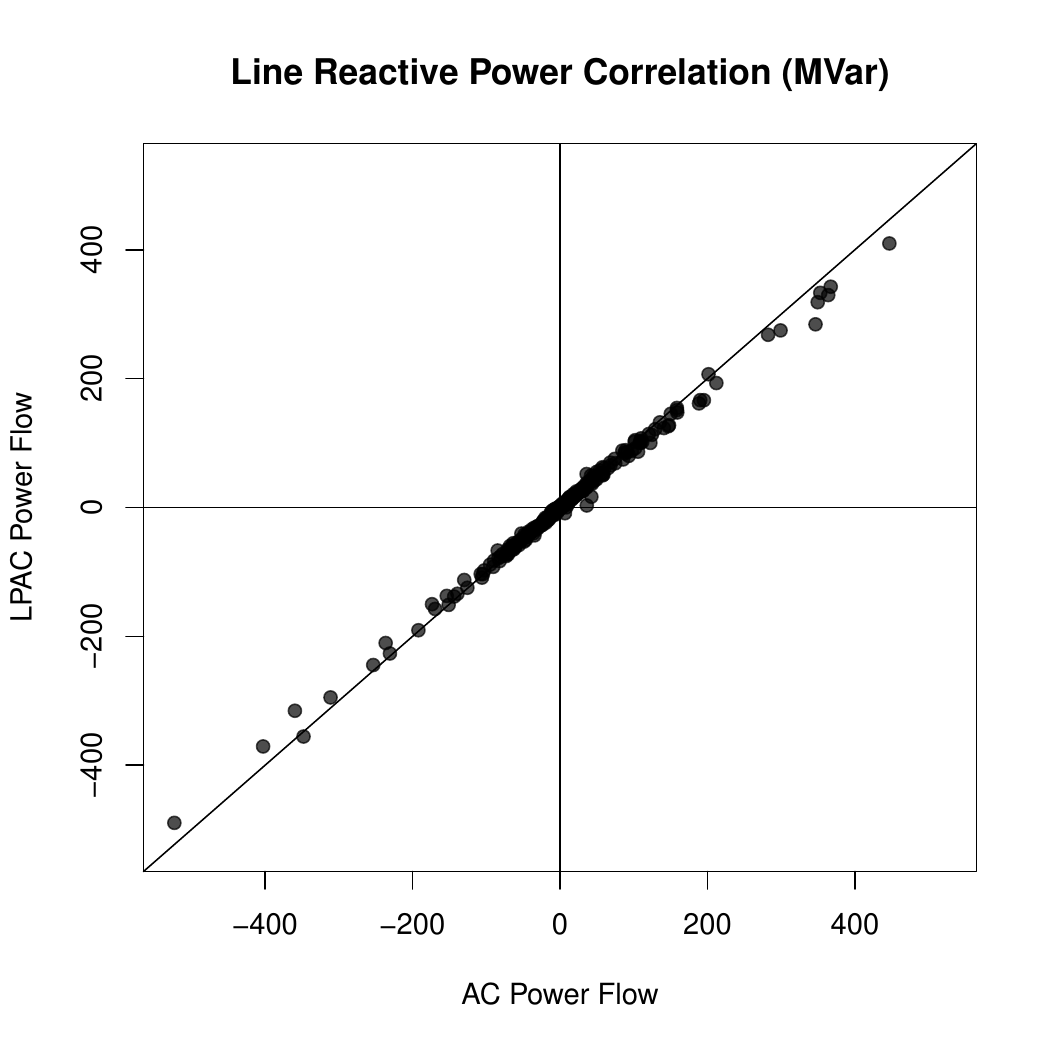} 
    \vspace{-0.6cm}
    \caption{Reactive Power Flow Correlation for the LPAC Model on IEEEdd17m (left) and MP300 (right) in a cold-start context.}
\label{fig:reactiveCorr}
\end{figure}

Table \ref{tbl:ldcs_reactive} presents the aggregate statistics on
line reactive power flows.  They indicate that reactive power flows
are generally accurate and highly precise in warm-start contexts.  To
highlight the model accuracy in cold-start contexts, the reactive flow
correlation for the two worst benchmarks, IEEEdd17m and MP300, is
presented in Figure \ref{fig:reactiveCorr}.

\begin{table}[t]
\center
\caption{Accuracy of the LPAC Model: Voltage Magnitudes.}
\begin{tabular}{|c||c||c|c|c|}
\hline
Benchmark & \multicolumn{4}{|c|}{Voltage Magnitude (Volts p.u.)} \\
 & Corr & $\mu(\Delta)$ & $\max(\Delta)$ & $\delta(\arg$-$\max(\Delta))$ \\
\hline
\multicolumn{5}{|c|}{The LPAC-Cold Model}\\

\hline
ieee14 & 0.9828 & 0.003524 & 0.01304 & 1.236 \\
\hline
mp24 & 0.9983 & 0.000676 & 0.003244 & 0.3362 \\
\hline
ieee30 & 0.9908 & 0.002445 & 0.01098 & 1.098 \\
\hline
mp30 & 0.9884 & 0.002186 & 0.009453 & 0.9723 \\
\hline
mp39 & 0.9992 & 0.0007521 & 0.002446 & 0.2313 \\
\hline
ieee57 & 0.9726 & 0.01038 & 0.03353 & 3.587 \\
\hline
ieee118 & 0.9989 & 0.000717 & 0.00476 & 0.4926 \\
\hline
ieeedd17 & 0.9651 & 0.01376 & 0.03345 & 3.424 \\
\hline
ieeedd17m & 0.9815 & 0.00647 & 0.01579 & 1.659 \\
\hline
mp300 & 0.9948 & 0.002361 & 0.01552 & 1.656 \\
\hline

\multicolumn{5}{|c|}{The LPAC-Warm Model}\\

\hline
ieee14 & 0.9998 & 0.0005479 & 0.001173 & 0.1111 \\
\hline
mp24 & 0.9996 & 0.000542 & 0.002214 & 0.2294 \\
\hline
ieee30 & 0.9994 & 0.001426 & 0.002508 & 0.25 \\
\hline
mp30 & 1.0000 & 0.0003884 & 0.000707 & 0.073 \\
\hline
mp39 & 0.9983 & 0.00154 & 0.003545 & 0.3524 \\
\hline
ieee57 & 0.9987 & 0.002138 & 0.005515 & 0.59 \\
\hline
ieee118 & 0.9999 & 0.0001961 & 0.001303 & 0.1344 \\
\hline
ieeedd17 & 0.9858 & 0.01204 & 0.02597 & 2.796 \\
\hline
ieeedd17m & 0.9760 & 0.009819 & 0.02037 & 2.067 \\
\hline
mp300 & 0.9967 & 0.002477 & 0.01403 & 1.52 \\
\hline

\end{tabular}
\label{tbl:ldcs_voltage}
\vspace{-0.2cm}
\end{table}

\begin{figure}[t]
\center
    \includegraphics[width=4.5cm]{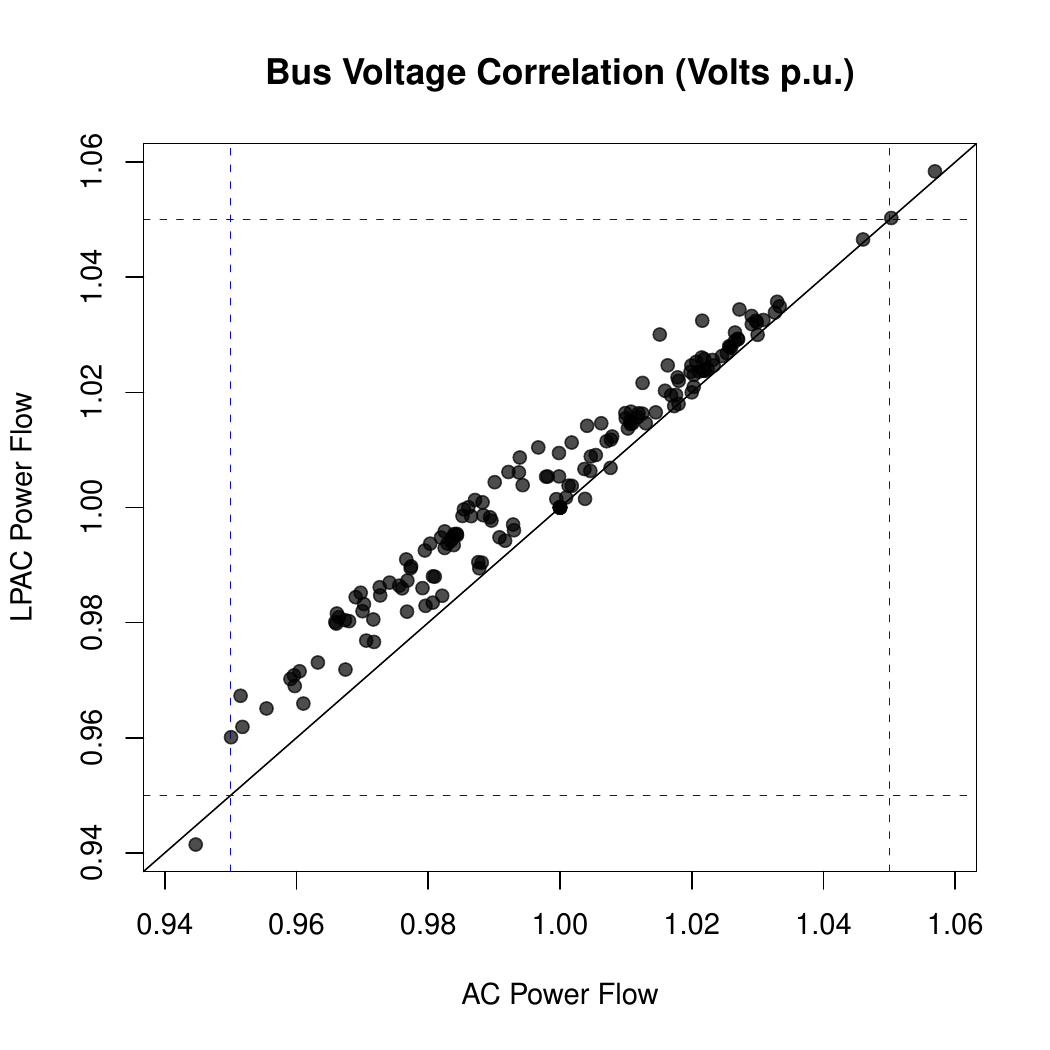}  
    \hspace{-0.4cm}
    \includegraphics[width=4.5cm]{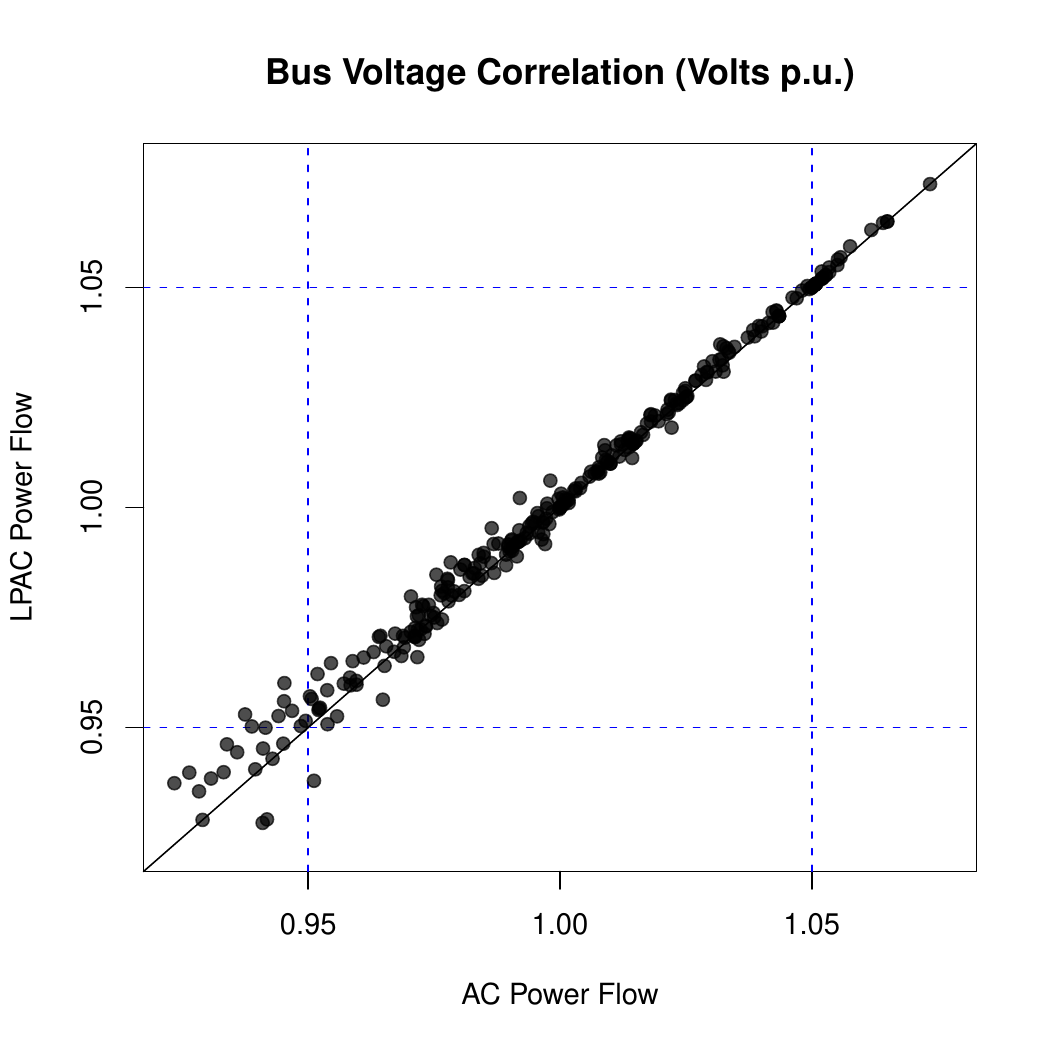} 
    \vspace{-0.6cm}
    \caption{Voltage magnitude correlation for Model LPAC on IEEEdd17m (left) and MP300 (right) in a cold-start context.}
\label{fig:voltageCorr}
\end{figure}

Table \ref{tbl:ldcs_voltage} presents the aggregate statistics on bus
voltage magnitudes.  These results indicate that voltage magnitudes
are very accurate on small benchmarks, but the accuracy reduces with
the size of the network.  The warm-start context brings a significant
increase in accuracy in larger benchmarks.  To illustrate the quality
of these solutions in cold-start contexts, the voltage magnitude
correlation for the two worst benchmarks, i.e., IEEEdd17m and MP300,
is presented in Figure \ref{fig:voltageCorr}.  The increase in voltage
errors is related to the distance from a load point to the nearest
generator.  The linearized voltage model incurs some small error on
each line.  As the voltage changes over many lines, these small errors
accumulate.  By comparing the percentage of voltage-controlled generator
buses in each benchmark $|G|/|N|$ (Table \ref{tbl:voltageControled}) to accuracy in Table
\ref{tbl:ldcs_voltage}, the IEEE57 and IEEEdd17 benchmarks indicate that that a
low percentage is a reasonable indicator of the voltage accuracy in the cold-start context.

\begin{table}[t]
\center

\caption{Percentage of Voltage-Controlled Buses in the Benchmarks.}
\begin{tabular}{|c|c|c|c|c|}
\hline
ieee14 & mp24 & ieee30 & mp30 & mp39 \\
\hline
35.7\% & 45.8\% & 20.0\% & 20.0\% & 30.8\%\\
\hline
\hline
ieee57 & ieee118 & ieeedd17 & mp300 & \\
\hline
12.3\% & 45.8\% & 7.4\% & 24.7\% & \\
\hline
\end{tabular}
\label{tbl:voltageControled}
\vspace{-0.2cm}
\end{table}

\subsection{Alternative Linear Models}
\label{section:exp:approx}

The formulation of the LPAC models explicitly removes two core
assumptions of the traditional LDC model:
\begin{enumerate}
\item Although
$\cos(\theta^\circ_n - \theta^\circ_m)$ maybe very close to $1$, those
small deviations are important.

\item Although $|g| \ll |b|$, the
conductance contributes significantly to the phase angles and voltage
magnitudes.
\end{enumerate}

\noindent
This section investigates three variants of the cold-start LPAC model that
reintegrate some of the assumptions of the LDC model.  The new models
are: (1) the LPAC-C model where only the cosine approximation is used
and $g=0$; (2) the LPAC-G model where only the $g$ value is used and
$cos(x)=1$; (3) the LPAC-CG model where $cos(x)=1$ and $g=0$. Tables
\ref{tbl:model_abs_err_var1} and \ref{tbl:model_abs_err_var2} present
the {\em cumulative absolute error} between the proposed linear
formulations and the true nonlinear solutions.  Many metrics may be of
interest but these results focus on line voltage drop $\widetilde{V}_n
- \widetilde{V}_m$ and bus power $\widetilde{S}_n$.  These were
selected because they are robust to errors which accumulate as power
flows through the network.  The results highlight two interesting
points.  First, all linear models tend to bring improvements over a
traditional LDC model.  Second, although integrating either the $g$
value or the cosine term brings some small improvement independently,
together they make significant improvements in accuracy.  Additionally
a comparison of Table \ref{tbl:model_abs_err_var1} and Table
\ref{tbl:model_abs_err_var2} reveals that the benefits of the new
linear models are more pronounced as the network increases.

\begin{table}[t]
\center
\caption{Accuracy Comparison of Various Linear Models (Part I).}
\begin{tabular}{|c||c|c|c|c|}
\hline

Model & \multicolumn{4}{|c|}{Cumulative Absolute Error} \\
& $\!\! \Re(\widetilde{V}_n\!\!-\!\widetilde{V}_m)\!\!$ & $\!\! \Im(\widetilde{V}_n\!\!-\!\widetilde{V}_m)\!\!$ & $ p_n$ & $ q_{n}$ \\

\hline
\multicolumn{5}{|c|}{ ieee14 }\\
\hline
LDC & 0.3839 & 0.166 & 13.39 & 118.4 \\
\hline
LPAC-A-GC & 0.1561 & 0.1296 & 13.39 & 140.7 \\
\hline
LPAC-A-G & 0.1221 & 0.1229 & 13.39 & 120.7 \\
\hline
LPAC-A-C & 0.1277 & 0.1262 & 8.843 & 53.24 \\
\hline
LPAC & 0.1008 & 0.1234 & 1.783 & 11.43 \\
\hline
\hline
\multicolumn{5}{|c|}{ mp24 }\\
\hline
LDC & 0.448 & 0.1434 & 53.22 & 792.4 \\
\hline
LPAC-A-GC & 0.3676 & 0.1289 & 53.22 & 546.2 \\
\hline
LPAC-A-G & 0.2309 & 0.1332 & 53.22 & 314.3 \\
\hline
LPAC-A-C & 0.2417 & 0.116 & 53.12 & 476.2 \\
\hline
LPAC & 0.03411 & 0.0828 & 6.94 & 64.12 \\
\hline
\hline
\multicolumn{5}{|c|}{ ieee30 }\\
\hline
LDC & 0.5429 & 0.2934 & 17.55 & 169.9 \\
\hline
LPAC-A-GC & 0.1607 & 0.2377 & 17.55 & 173.8 \\
\hline
LPAC-A-G & 0.1284 & 0.2268 & 17.55 & 147.7 \\
\hline
LPAC-A-C & 0.1587 & 0.1638 & 15.99 & 66.28 \\
\hline
LPAC & 0.1305 & 0.1476 & 2.9 & 16.72 \\
\hline
\hline
\multicolumn{5}{|c|}{ mp30 }\\
\hline
LDC & 0.4341 & 0.1728 & 2.444 & 181 \\
\hline
LPAC-A-GC & 0.166 & 0.1584 & 2.444 & 20.78 \\
\hline
LPAC-A-G & 0.1616 & 0.1581 & 2.444 & 17.96 \\
\hline
LPAC-A-C & 0.06886 & 0.1454 & 2.444 & 9.387 \\
\hline
LPAC & 0.07338 & 0.1456 & 1.736 & 6.76 \\
\hline
\hline
\multicolumn{5}{|c|}{ mp39 }\\
\hline
LDC & 0.5634 & 0.1997 & 43.64 & 2816 \\
\hline
LPAC-A-GC & 0.2449 & 0.2105 & 43.64 & 950.3 \\
\hline
LPAC-A-G & 0.1418 & 0.1958 & 43.64 & 284.1 \\
\hline
LPAC-A-C & 0.1896 & 0.1868 & 37.65 & 947.9 \\
\hline
LPAC & 0.04462 & 0.1337 & 0.4745 & 93.31 \\
\hline

\end{tabular}
\label{tbl:model_abs_err_var1}
\vspace{-0.2cm}
\end{table}

\begin{table}[t]
\center
\caption{Accuracy Comparison of Various Linear Models (Part II).}
\begin{tabular}{|c||c|c|c|c|}
\hline

Model & \multicolumn{4}{|c|}{Cumulative Absolute Error} \\
 & $\!\! \Re(\widetilde{V}_n\!\!-\!\widetilde{V}_m)\!\!$ & $\!\! \Im(\widetilde{V}_n\!\!-\!\widetilde{V}_m)\!\!$ & $ p_n$ & $ q_{n}$ \\

\hline
\hline
\multicolumn{5}{|c|}{ ieee57 }\\
\hline
LDC & 1.343 & 0.6803 & 27.9 & 529.1 \\
\hline
LPAC-A-GC & 0.4158 & 0.5489 & 27.9 & 264 \\
\hline
LPAC-A-G & 0.3238 & 0.5311 & 27.9 & 264.1 \\
\hline
LPAC-A-C & 0.3773 & 0.4153 & 24.43 & 113.8 \\
\hline
LPAC & 0.3647 & 0.3854 & 4.736 & 26.35 \\
\hline
\hline
\multicolumn{5}{|c|}{ ieee118 }\\
\hline
LDC & 3.083 & 1.239 & 132.7 & 2152 \\
\hline
LPAC-A-GC & 0.7298 & 0.9944 & 132.7 & 1364 \\
\hline
LPAC-A-G & 0.6502 & 0.9929 & 132.7 & 1194 \\
\hline
LPAC-A-C & 0.4417 & 0.8553 & 104.3 & 750 \\
\hline
LPAC & 0.2625 & 0.5252 & 0.7279 & 142.1 \\
\hline
\hline
\multicolumn{5}{|c|}{ ieeedd17 }\\
\hline
LDC & 4.144 & 3.263 & 201.3 & 3857 \\
\hline
LPAC-A-GC & 5.783 & 4.881 & 201.3 & 2719 \\
\hline
LPAC-A-G & 4.169 & 3.242 & 201.3 & 616.4 \\
\hline
LPAC-A-C & 4.162 & 3.48 & 200.5 & 2660 \\
\hline
LPAC & 1.135 & 1.019 & 30.38 & 362.1 \\
\hline
\hline
\multicolumn{5}{|c|}{ ieeedd17m }\\
\hline
LDC & 3.798 & 1.972 & 191.1 & 3353 \\
\hline
LPAC-A-GC & 5.152 & 3.219 & 191.1 & 2210 \\
\hline
LPAC-A-G & 3.302 & 2.111 & 191.1 & 389.4 \\
\hline
LPAC-A-C & 3.31 & 2.118 & 190.4 & 2146 \\
\hline
LPAC & 0.49 & 0.6324 & 20.21 & 223.7 \\
\hline
\hline
\multicolumn{5}{|c|}{ mp300 }\\
\hline
LDC & 13.76 & 4.689 & 418.5 & 14240 \\
\hline
LPAC-A-GC & 11.2 & 5.324 & 418.5 & 5595 \\
\hline
LPAC-A-G & 9.831 & 5.171 & 418.5 & 1648 \\
\hline
LPAC-A-C & 7.855 & 4.403 & 348.5 & 5434 \\
\hline
LPAC & 0.8699 & 1.378 & 9.703 & 976.4 \\
\hline

\end{tabular}
\label{tbl:model_abs_err_var2}
\vspace{-0.2cm}
\end{table}

\section{Case Studies}
\label{section:casestudies}

This section describes two case studies to evaluate the potential of
the LPAC models: Power restoration and capacitor placement. The goal
is not to present comprehensive solutions for these two complex
problems, but to provide preliminary evidence that the LPAC models may
be useful in striking a good compromise between efficiency and
accuracy for such applications. This section should be
viewed as presenting a ``proof-of-concept" that the LPAC models may be
valuable for certain classes of applications where the LDC model is
not accurate enough and existing approaches are too time consuming or
suboptimal.

\subsection{Power Restoration}
\label{section-results-restoration}

After a significant disruption due to, say, a natural disaster, large
sections of the power network need to be re-energized.  To understand
the effects of restoration actions, power engineers must simulate the
network behaviour under various courses of action.  However, the
network is far from its normal operating state, which makes it
extremely challenging to solve the AC power flow equations.  In fact,
the task of finding an AC solution without a reasonable starting point
has been regarded as "maddeningly difficult" \cite{Overbye:2004vb}.
The LPAC model studied here has the benefit of providing starting
values for all the variables in the AC power flow problem, unlike the
traditional LDC which only provides active power values. Furthermore,
the LPAC model has the additional advantage of supporting bounds on
reactive generation and voltage magnitudes and such constraints are
critical for providing feasible solutions to the AC power flow. This
section illustrates these benefits.

Before presenting the power-restoration model, it is important to
mention the key aspect of this application.  When the power system
undergoes significant damages, load shedding must occur.  The LDC and
LPAC models must be embedded in a restoration model that maximizes the
served load given operational constraints such as the generation
limits.  These load values indicate the maximum amount of power that
can be dispatched while ensuring system stability.  Model
\ref{model:linearRecovery} presents a linear program based on the
warm-start LPAC model which, given limits on active power generation
$\overline{p^g}$ and the desired active and reactive loads
$\overline{p^l}, \overline{q^l}$ at each bus, determines the maximum
amount of load that can be dispatched.  The model assumes that the
loads can be shed continuously and that the active and reactive parts
of the load should maintain the same power factor.  The objective
function (M\ref{model:linearRecovery}.1) maximizes the percentage of
served load. Constraints (M\ref{model:linearRecovery}.2) and
(M\ref{model:linearRecovery}.3) set the active and reactive injection
at bus $n$ appropriately based on the decision variables for load
shedding and generation dispatch. Constraint
(M\ref{model:linearRecovery}.4) ensures that reactive generation only
occurs at generator buses and Constraint
(M\ref{model:linearRecovery}.5) now defines $q_n$ for generator buses
as well.

Since it reasons about reactive power and voltage magnitudes, Model
\ref{model:linearRecovery} can be further enhanced to impose bounds
on these values. As we will show, such bounds are often critical to
obtain high-quality solutions in power restoration contexts. If a
reactive generation bound $\overline{q^g}$ is supplied, this model can
be extended by adding the constraint,
\[
q^g_n \leq \overline{q^g_n}  \;\; \forall n \in N.
\]
Voltage magnitude limits can also be incorporated.  Given upper and
lower voltage limits $\overline{|\widetilde{V}|}$ and 
$\underline{|\widetilde{V}|}$, the constraint 
\[
\underline{|\widetilde{V}|} \leq 1.0 + \phi_{n} \leq \overline{|\widetilde{V}|}  \;\; \forall n \in N.
\]
may be used to enforce bounds on voltage magnitudes. The experimental
results study the benefits of the LPAC model, suitably enhanced to
capture these extensions, for power restoration. They compare a
variety of linear models including the LDC model, the LPAC model, and
enhancements of the LPAC model with additional constraints on reactive
power and voltage magnitudes.

\begin{model}[t]
\footnotesize
\begin{tabbing}
123\= $\theta_{i} \in (-\infty, \infty)$1 \= - phase angle on bus $i$ (radians)\=\kill
{\bf Inputs:}\\	
\> $\overline{p^g_n}$     \> - maximum active injection for bus $n$\\
\> $\overline{p^l_n}$     \> - desired active load at bus $n$\\
\> $\overline{q^l_n}$     \> - desired reactive load at bus $n$\\
\> Inputs from Model \ref{model:LPACwarm} (The Warm-Start LPAC Model) \\
{\bf Variables:}\\
\> $p^g_n \in (0, \overline{p^g_n})$ \> - active generation at bus  $n$\\
\> $q^g_n \in (-\infty, \infty)$ \> - reactive generation at bus $n$\\
\> $l_n \in (0, 1)$ \> - percentage of load served at bus $n$\\
\> Variables from Model \ref{model:LPACwarm} (The Warm-Start LPAC Model) \\
{\bf Maximize:}\\
123\= $\underline{|\widetilde{V}|} \leq 1.0 + \phi_{n} \leq 1.05 \;\; \forall n \in N$  123\=\=(2)\kill
\> $\displaystyle \sum_{n \in N} l_n$ \>\> (M\ref{model:linearRecovery}.1) \\
{\bf Subject to:}\\
\> $p_{n}  = -\overline{p^l_{n}}l_n + p^g_n \;\; \forall n \in N$ \>\> (M\ref{model:linearRecovery}.2)\\
\> $q_{n}  = -\overline{q^l_{n}}l_n + q^g_n \;\; \forall n \in N$ \>\> (M\ref{model:linearRecovery}.3)\\
\> $q^g_{n} = 0 \;\; \forall n \in N \setminus G$ \>\> (M\ref{model:linearRecovery}.4)\\
\> $q_{n} = \displaystyle \sum_{m \in N}^{n \neq m} \hat{q}^t_{nm} + \hat{q}^{\Delta}_{nm} \;\; \forall n \in G$ \>\> (M\ref{model:linearRecovery}.5)\\
\> Constraints from Model \ref{model:LPACwarm} (The Warm-Start LPAC Model)
\end{tabbing}
\caption{A LP for Maximizing Desired Load.}
\label{model:linearRecovery}
\end{model}

\begin{table}[t]
\center
\caption{Power Restoration: Achieving AC Feasibility From Different Models.}
\begin{tabular}{|c||c|c|c|c|}
\hline
Scenario & LDC & LPAC & LPAC-R & LPAC-R-V \\
\hline
$N\!\!-\!3$ & 998 &  999 & 1000 &  1000  \\
\hline
$N\!\!-\!4$ & 999 & 1000 & 1000  & 1000  \\
\hline
$N\!\!-\!5$ & 987  & 994 & 1000  & 1000\\
\hline
$N\!\!-\!6$ & 507  & 594 & 903 &  1000\\
\hline
$N\!\!-\!7$ & 738  & 856 & 973  & 974  \\
\hline
$N\!\!-\!8$ & 949  & 996 & 1000 &1000  \\
\hline
$N\!\!-\!9$ & 847  & 932 & 1000  &1000  \\
\hline
$N\!\!-\!10$ & 219 & 452 & 992  & 999 \\
\hline
$N\!\!-\!11$ & 726  & 972 & 1000 & 997 \\
\hline
$N\!\!-\!12$ & 491  &  779 & 998  & 999 \\
\hline
$N\!\!-\!13$ & 444  &  617 & 983 & 991 \\
\hline
$N\!\!-\!14$ & 545  & 637 & 998 & 1000\\
\hline
$N\!\!-\!15$ & 1000  & 1000  & 1000  & 1000 \\
\hline
$N\!\!-\!16$ & 989   & 1000 & 1000  &1000  \\
\hline
$N\!\!-\!17$ & 1000 & 1000 & 1000  &  1000\\
\hline
$N\!\!-\!18$ & 969  & 1000 & 1000 &1000  \\
\hline
$N\!\!-\!19$ & 999  & 1000  &  1000 & 1000 \\
\hline
$N\!\!-\!20$ & 1000 & 1000 & 1000  &  1000\\
\hline
\end{tabular}
\label{tbl:restoration}
\end{table}

\begin{table}[t]
\center
\caption{Power Restoration: Average Load Shedding (\% of Total Active Power).}
\begin{tabular}{|c||c|c|c|c|}
\hline
Scenario & LDC & LPAC & LPAC-R & LPAC-R-V \\
\hline
$N\!\!-\!3$ & 3.23 & 6.193 & 15.94 & 15.99 \\
\hline
$N\!\!-\!4$ & 2.827 & 3.55 &  9.183 & 9.278 \\
\hline
$N\!\!-\!5$ & 0.9562 & 2.204 &  8.027 & 8.082 \\
\hline
$N\!\!-\!6$ & 5.805 & 6.149 &  9.287  & 9.921 \\
\hline
$N\!\!-\!7$ & 1.506 & 5.709 &  19.43 & 19.46 \\
\hline
$N\!\!-\!8$ & 13.78 & 18.54 &  27.37 & 27.39 \\
\hline
$N\!\!-\!9$ & 15.92 & 24.27 &  42.76 & 42.78 \\
\hline
$N\!\!-\!10$ & 29.23 & 24.02 &  32.09 & 32.25 \\
\hline
$N\!\!-\!11$ & 25.18 & 25.24 &  42.62 & 42.63 \\
\hline
$N\!\!-\!12$ & 36.35 & 28.25 &  38.03 & 38.82 \\
\hline
$N\!\!-\!13$ & 40.1 & 32.56 &  38.3 & 38.66 \\
\hline
$N\!\!-\!14$ & 39.98 & 36.9 &  40.45 & 40.67 \\
\hline
$N\!\!-\!15$ & 81.91 & 81.92 &  81.92 & 81.92 \\
\hline
$N\!\!-\!16$ & 86.21 & 86.31 &  86.32 & 86.32 \\
\hline
$N\!\!-\!17$ & 89.89 & 89.89 &  89.89 & 89.89 \\
\hline
$N\!\!-\!18$ & 88.26 & 88.3 &  88.32 & 88.32 \\
\hline
$N\!\!-\!19$ & 85.9 & 86.13 &  86.13 & 86.13 \\
\hline
$N\!\!-\!20$ & 86.2 & 86.37 &  86.38 & 86.38 \\
\hline
\end{tabular}
\label{tbl:restoration-shedding}
\end{table}

Table \ref{tbl:restoration} studies the applicability of various
linear power models for network restoration on the IEEE30 benchmark.
1000 line outage cases were randomly sampled from each of the
$N\!-\!3, N\!-\!4, N\!-\!5,\ldots,N\!-\!20$ contingencies.  Each
contingency is solved with a linear power model (e.g., the LDC model
or the LPAC model), whose solution is used as a starting point for the
AC model. The performance metric is the number of cases where the AC
solver converges, as a good linear model should yield a feasible
generation dispatch with a good starting point for the AC solver. To
understand the importance of various network constraints, four linear
models are studied: the traditional LDC model; the LPAC model; the
LPAC model with constraints on reactive generation (LPAC-R); and the
LPAC with constraints on reactive generation and voltage limits
(LPAC-R-V).  The number of solved models for each of the contingency
classes is presented in Table \ref{tbl:restoration}.  The results
indicate that a traditional LDC model is overly optimistic and often
produces power dispatches that do not lead to feasible AC power flows
(the $N\!\!-\!10$ and $N\!\!-\!13$ are particularly striking).
However, each refinement of the LPAC model solves more contingencies.
The LPAC-R-V model is very reliable and is able to produce feasible
dispatches in all contingencies except 40 . 
This means that the LPAC-R-V model solves 99.76\% of the 17,000
contingencies studied. Table \ref{tbl:restoration-shedding} depicts
the load shed by the various models. For large contingencies, the
LPAC-R-V model not only provides good starting points for an AC solver
but its load shedding is only slightly larger than the (overly
optimistic) LDC model. These results provide compelling evidence of
the benefits of the LPAC model for applications dealing with
situations outside the normal operating conditions. In addition, Model
\ref{model:linearRecovery} can replace the LDC model in power
restoration applications (e.g., \cite{SSP1,PRVRP1}) that are using MIP
models to minimize the size of a blackout over time.

\subsection{The Capacitor Placement Problem}

The Capacitor Placement Problem (CPP) is a well-studied application
\cite{Aguiar:2005pscc,871731,544656} and many variants of the problem
exist. This section uses a simple version of the problem to
demonstrate how the LPAC model can be used as a building block inside
a MIP solver for decision-support applications. 

Informally speaking, the CPP consists of placing capacitors throughout
a power network to improve voltage stability.  The version studied
here aims at placing as few capacitors as possible throughout the
network, while meeting a lower bound $\underline{|\widetilde{V}|}$ on
the voltages and satisfying a capacitor injection limit
$\overline{q^c}$ and reactive generation limits $\overline{q^g_n} \;\;
(n \in G)$.  Model \ref{model:LPACcoldCap} presents a CPP model based
on the {\em cold-start} LPAC model.  For each bus $n$, the additional
decision variables are the amount of reactive support added by the
capacitor $q^c_n$ and a variable $c_n$ indicating whether a capacitor
was used.

The objective function (M\ref{model:LPACcoldCap}.1) minimizes the
number of capacitors.  Constraints (M\ref{model:LPACcoldCap}.2) ensure
the voltages do not drop below the desired limit and do not exceed the
preferred operating condition of 1.05 Volts p.u.  Constraints
(M\ref{model:LPACcoldCap}.3) link the capacitor injection variables
with the indicator variables, a standard technique in MIP models.
Constraints (M\ref{model:LPACcoldCap}.4) ensures each generator $n \in
G$ does not exceed its reactive generation limit $\overline{q^g_n}$.
and constraints (M\ref{model:LPACcoldCap}.5) defines the reactive
power for generators. Lastly, Constraints (M\ref{model:LPACcoldCap}.6)
redefines the reactive power equation to inject the capacitor
contribution $q^c$. The remainder of the model is the same as Model
\ref{model:LPACcold} (the cold-start LPAC model).

The CPP model was tested on a modified version of the IEEE57
benchmark.  All of the IEEE benchmarks have sufficient reactive
support in their normal state.  To make an interesting capacitor
placement problem, the transformer tap ratios are set to 1.0 and
existing synchronous condensers are removed.  This modified benchmark
(IEEE57-C) has significant voltage problems with several bus voltages
dropping below 0.9.  By design, a solution to Model
\ref{model:LPACcoldCap} satisfies all of the desired constraints.
However, Model \ref{model:LPACcoldCap} is based on the LPAC model and
is only an approximation of the AC power flow.  To understand the {\em
  true} value of Model \ref{model:LPACcoldCap}, we solve the resulting
solution network with an AC solver and measure how much the
constraints are violated.  Table \ref{tbl:capp_sol} presents the
results of Model \ref{model:LPACcoldCap} on benchmark IEEE57-C with
$\overline{q^c} = 30$ and various thresholds
$\underline{|\widetilde{V}|}$. The table presents the following
quantities: The minimum desired voltage $\underline{|\widetilde{V}|}$;
The worst violation of the voltage lower-bound
$\min(|\widetilde{V}|)$; The worst violation of the voltage upper
bound $\max(|\widetilde{V}|)$; The worst violation of reactive
injection upper bound $\max(q_n)$; The number of capacitors placed
$\sum c_n$; and the runtime of the MIP to prove the optimal placement
solution.  The table indicates that the CPP model is extremely
accurate and only has minor constraint violations on the lower bounds
of the voltage values.  It is important to note that, although the CPP
model can take as long as five minutes to prove optimality\footnote{It
  is of course only optimal up the quality of the LPAC
  approximation.}, it often finds the best solution value within a few
seconds.  The voltage lower bound approaches the value of $0.985$, 
which is the lowest value of the voltage-controlled generators in the
benchmark. These results remain consistent for other voltage bounds.

Once again, the CPP model indicates the benefits of the LPAC
approximation for decision-support applications that need to reason
about reactive power and voltages.

\begin{model}[t]
\footnotesize
\begin{tabbing}
123\= $\theta_{i} \in (-\infty, \infty)$1 \= - phase angle on bus $i$ (radians)\=\kill
{\bf Inputs:}\\	
\> $\overline{q^g_n}$     \> - injection bound for generator $n$\\
\> $\overline{q^c}$     \> - capacitor injection bound\\
\> $\underline{|\widetilde{V}|}$     \> - minimum desired voltage magnitude\\
\> Inputs from Model \ref{model:LPACcold} (The Cold-Start LPAC Model) \\
{\bf Variables:}\\
\> $q^c_n \in (0, \overline{q^c})$ \> - capacitor reactive injection\\
\> $c_n \in \{0, 1\}$ \> - capacitor placement indicator\\
\> Variables from Model \ref{model:LPACcold} (The Cold-Start LPAC Model)  \\
{\bf Minimize:}\\
\> $q_{n} - q^c_n = \displaystyle \sum_{m \in N}^{n \neq m} \hat{q}^t_{nm} + \hat{q}^{\Delta}_{nm} \;\; \forall n \in N: n \neq s \wedge n \not\in G$ 123\=\= \kill
\> $\displaystyle \sum_{n \in N} c_n$ \>\> (M\ref{model:LPACcoldCap}.1) \\
{\bf Subject to:}\\
\> $\underline{|\widetilde{V}|} \leq 1.0 + \phi_{n} \leq 1.05 \;\; \forall n \in N$ \>\> (M\ref{model:LPACcoldCap}.2)\\
\> $q^c_n \leq Mc_n$ \>\> (M\ref{model:LPACcoldCap}.3)\\
\> $q_{n}  \leq \overline{q^g_n} \;\; \forall n \in G$ \>\> (M\ref{model:LPACcoldCap}.4)\\
\> $q_{n} = \displaystyle \sum_{m \in N}^{n \neq m} \hat{q}^t_{nm} + \hat{q}^{\Delta}_{nm} \;\; \forall n \in G$ \>\> (M\ref{model:LPACcoldCap}.5)\\
\> $q_{n} - q^c_n = \displaystyle \sum_{m \in N}^{n \neq m} \hat{q}^t_{nm} + \hat{q}^{\Delta}_{nm} \;\; \forall n \in N: n \neq s \wedge n \not\in G$ \>\> (M\ref{model:LPACcoldCap}.6)\\
\> Constraints from Model \ref{model:LPACcold}  (The Cold-Start LPAC Model) except (M\ref{model:LPACcold}.5)
\end{tabbing}
\caption{A MIP for the Capacitor Placement Problem.}
\label{model:LPACcoldCap}
\end{model}

\begin{table}[t]
\center
\caption{Capacitor Placement: Effects of $\underline{|\widetilde{V}|}$ on IEEE57-C, $\overline{q^c} = 30$ MVar}
\begin{tabular}{|c||c|c|c|c|c|c|}
\hline
$\underline{|\widetilde{V}|}$ & $\min(|\widetilde{V}|)$&  $\max(|\widetilde{V}|)$ & $\max(q_n)$& $\sum c_n$ &  Time (sec.) \\
\hline
0.8850 &  0.000000 & 0.0  & 0.0 & 1  & 1 \\
\hline
0.9350 &  0.000000 & 0.0  & 0.0 & 3  & 8 \\	
\hline
0.9600 & 0.000000 & 0.0  & 0.0 & 5 & 156 \\
\hline
0.9750 &  0.000000 & 0.0  & 0.0 & 6 & 177 \\
\hline
0.9775 & 0.000000 & 0.0  & 0.0 & 6  & 139 \\
\hline
0.9800 & 0.000000 & 0.0  & 0.0 & 6  & 75 \\ 
\hline
0.9840 & -0.000802 & 0.0 & 0.0 & 7 & 340 \\ 
\hline
\end{tabular}
\label{tbl:capp_sol}
\vspace{-0.3cm}
\end{table}

\section{Related Work}
\label{section:Related}

Many linearizations of the AC power flow equations have been developed
\cite{Wang1996153,Thukaram198472,5971792,IJIESP2009,DBLP:conf/inoc/KosterL11,5772044,4075431}.
Broadly, they can be grouped into iterative methods
\cite{Thukaram198472,IJIESP2009,4075431} and convex models
\cite{Wang1996153,5971792,DBLP:conf/inoc/KosterL11,5772044}.

\paragraph*{Iterative Methods} Iterative methods, such as the fast-decoupled
load flow \cite{4075431}, significantly reduce the computation time of
solving the AC equations and demonstrate sufficient accuracy. Their
disadvantage however is that they cannot be efficiently integrated
into traditional decision-support tools. Indeed, MIP solvers require
purely declarative models to obtain lower bounds that are critical in
reducing the size of the search space. Note however that, modulo the
linear approximations, the LPAC model can be viewed as solving a
decoupled load flow globally.  The key differences are:
\begin{enumerate}
\item Because the model forms one large linear system, all of the
  steps of the decoupled load flow are effectively solved
  simultaneously; 

\item Because the formulation is a linear program, the values of $p$
  and $q$ can now be decision variables, and bounds may be placed on the
  line capacities, voltage magnitudes, and phase angles;

\item The model may be embedded in a MIP solver for making discrete
  decisions about the power system.
\end{enumerate}
The second and third points represent significant advantages over the fast-decoupled
load flow and other iterative methods.

\paragraph*{Convex Models} Although many variants of the LDC model
exist, few declarative models incorporate reactive flows in cold-start
contexts.  To our knowledge, three cold-start approaches have been
proposed: (1) a polynomial approximation scheme \cite{5772044}, (2) a
semi-definite programming relaxation \cite{5971792}, and (3) a
voltage-difference model \cite{DBLP:conf/inoc/KosterL11}.

The polynomial approximation has the advantage of solving a convex
relaxation of the AC power equations but the number of variables and
constraints needed to model the relaxation "grows rapidly"
\cite{5772044} and only second-order terms were considered.  The
accuracy of this approach for general power flows remains an open
question: Reference \cite{5772044} focuses on a transmission planning
application and does not quantify the accuracy of the approximation
relative to an AC power flow.

The semi-definite programming (SDP) relaxation \cite{5971792} has the
great advantage that it can solve the power flow equations precisely,
without any approximation.  In fact, reference \cite{5971792}
demonstrated that the formulation finds the globally optimal value to
the AC optimal power flow problem on a number of traditional
benchmarks.  However, recent work has shown this does not hold on some
practical examples \cite{6120344}.  Computationally, SDP solvers are
also less mature than LP solvers and their scalability remains an open
question \cite{SDPBench}.  Solvers integrating discrete variables on
top of SDP models \cite{1393890} are very recent and do not have the
scientific maturity of MIP solvers \cite{Bixby2000}.

The voltage-difference model \cite{DBLP:conf/inoc/KosterL11} has a
resemblance to a model combining the equation
\begin{eqnarray*}
  \hat{p}^h_{nm} &=& |\widetilde{V}^h_n|^{2}g_{nm} - |\widetilde{V}^h_n||\widetilde{V}^h_m|g_{nm} 
  -|\widetilde{V}^h_n||\widetilde{V}^h_m|b_{nm}(\theta^\circ_{n} - \theta^\circ_{m}) \label{eq:acActiveHSCos1}
\end{eqnarray*}
with Equation (\ref{eq:acReactiveLineColdApproxFinal}).  However, it
makes a fundamental assumption that all voltages are the same before
computing the voltage differences.  In practice, voltage-controlled
generators violate this assumption.  On traditional power system
benchmarks, we observed that the voltage-difference formulation had
similar accuracy to the LDC model.

\section{Conclusion}
\label{section:conclusion}

This paper presented linear programs to approximate the AC power flow
equations. These linear programs, called the LPAC models, capture both
the voltage phase angles and magnitudes, which are coupled through
equations for active and reactive power. The models use a piecewise
linear approximation of the cosine term in the power flow
equations. The cold-start and warm-start models use a Taylor series
for approximating the remaining nonlinear terms.

The LPAC models have been evaluated experimentally over a number of
standard benchmarks under normal operating conditions and under
contingencies of various sizes. Experimental comparisons with AC
solutions on a variety of standard IEEE and MatPower benchmarks shows
that the LPAC models are highly accurate for active and reactive
power, phase angles, and voltage magnitudes. The paper also presented
two case studies in power restoration and capacitor placement to
provide evidence that the cold-start and wam-start LPAC models can be
efficiently used as a building block for optimization problems
involving constraints on reactive power flow and voltage
magnitudes. As a result, the LPAC models have the potential to broaden
the success of the traditional LDC model into new application areas
and to bring increased accuracy and reliability to current LDC
applications.

There are many opportunities for further study, including the
application of the LPAC models to a number of application areas.  From
an analysis standpoint, it would be interesting to compare the LPAC
models with an AC solver using a ``distributed slack bus''. Such an AC
solver models the real power systems more accurately and provides a
better basis for comparison, since the LPAC models are easily extended
to flexible load and generation at all buses. 


\appendices
\section{A Linear Programming Approximation of Cosine}
\label{section:lpcos}

\begin{figure}[t]
  \centering
  \includegraphics[width=7.5cm]{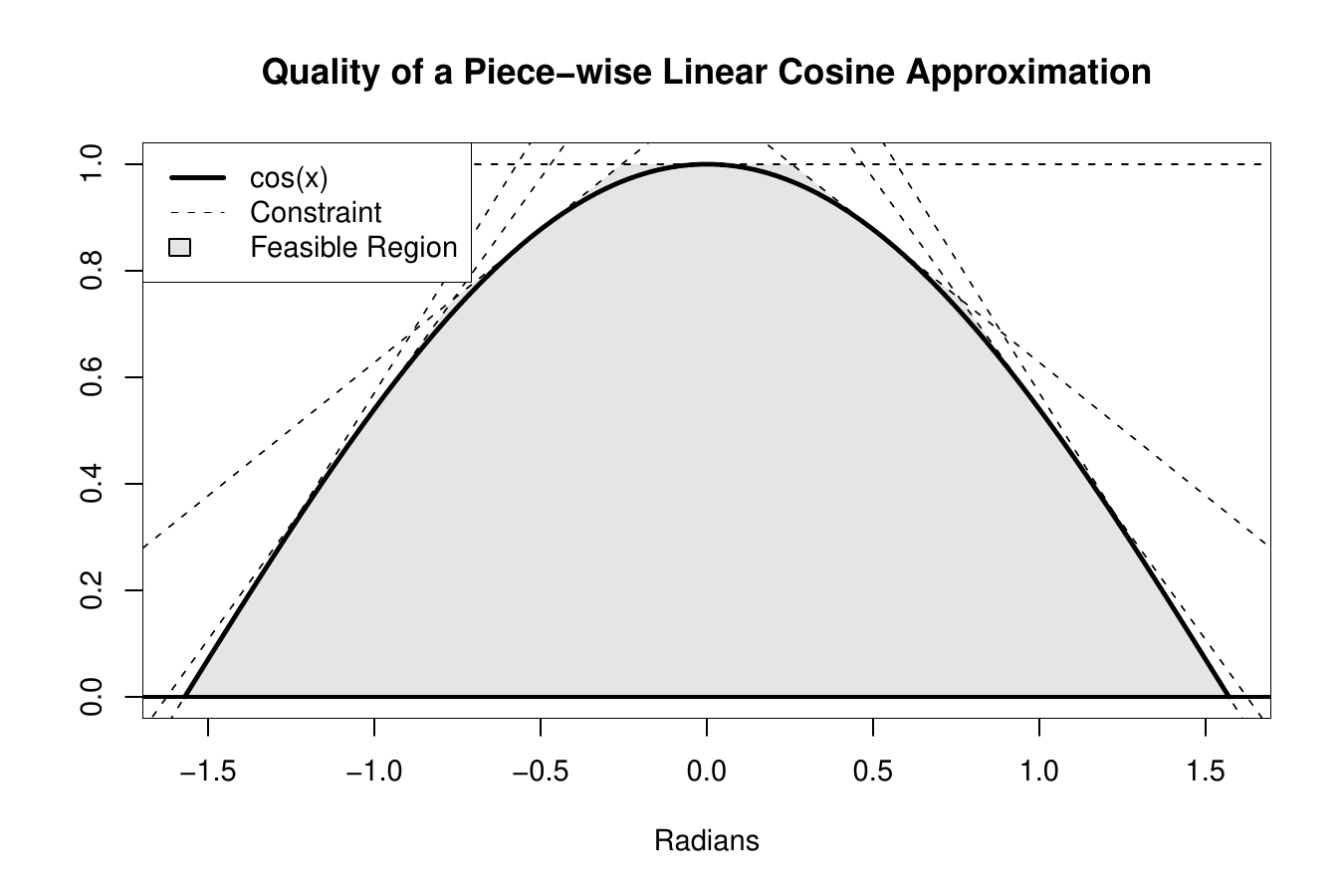} 
  \vspace{-0.5cm}
  \caption{A Piecewise-Linear Approximation of Cosine using 7 Inequalities.}
\label{fig:cos_approx}
\end{figure}

\noindent
The convex approximation of the cosine function is implemented through a
  piecewise linear function that produces a linear program in the
  following way.  The modeler selects a desired domain
  $(l,h)$\footnote{The domain should not exceed the range
      $(-\pi/2,\pi/2)$ to ensure convexity. In practice,
      $\theta^\circ_{n} - \theta^\circ_{m}$ is typically very small
      and a narrower domain is preferable.}  and a number segments
  $s$.  Then $s$ tangent inequalities are placed on the cosine
  function within the provided domain to approximate the convex
  region. Figure \ref{fig:cos_approx} illustrates the approximation
approach using seven linear inequalities.  The dark black line shows
the cosine function, the dashed lines are the linear inequality
constraints, and the shaded area is the feasible region of the linear
system formed by those constraints. The inequalities are obtained from
tangents lines at various points on the function.  Specifically, for an 
x-coordinate $a$, the tangent line is $y = -\sin(a)(x - a) + \cos(a)$ and,
within the domain of $(-\pi/2,\pi/2)$, the inequality
\[
y \leq -\sin(a)(x - a) + \cos(a) \;\; \forall a
\] 
holds. Figure \ref{fig:pwl_cos} gives an algorithm to generate $s$
inequalities evenly spaced within $(l,h)$. In the algorithm, $x$ is a
decision variable used as an argument of the cosine function and
$x_{\widehat{\cos}}$ is a decision variable capturing the approximate
value of $cos(x)$. 

\begin{figure}[t]
\center
\begin{minipage}{0cm}
\begin{algorithm}{PWL$<$COS$>$}{x_{\widehat{\cos}}, x, l, h, s}
post(x_{\widehat{\cos}} \geq \frac{\cos(h) - \cos(l)}{h-l}(x - l) + \cos(l))\\
inc \= (h-l)/(s+1)\\
a \= l + inc\\
\begin{FOR}{i \in 1..s}
  f_{a} \= \cos(a)\\
  s_{a} \= -\sin(a)\\
  post(x_{\widehat{\cos}} \leq s_{a}x - s_{a}a + f_{a})\\
  a \= a + inc
\end{FOR}
\end{algorithm}
\end{minipage}
\vspace{-0.5cm}
\caption{Generating Evenly Spaced Piecewise-Linear Approximations.}
\label{fig:pwl_cos}
\vspace{-0.5cm}
\end{figure}

\bibliographystyle{IEEEtran}
\bibliography{IEEEabrv,../../SIM}

%
%



\end{document}